\documentclass{article}

\PassOptionsToPackage{numbers}{natbib}

\usepackage[final]{nips_2018}

\usepackage[utf8]{inputenc} % allow utf-8 input
\usepackage[T1]{fontenc}    % use 8-bit T1 fonts
\usepackage{hyperref}       % hyperlinks
\usepackage{url}            % simple URL typesetting
\usepackage{booktabs}       % professional-quality tables
\usepackage{amsfonts}       % blackboard math symbols
\usepackage{nicefrac}       % compact symbols for 1/2, etc.
\usepackage{microtype}      % microtypography

\usepackage{comment}

\usepackage{graphicx}
\usepackage{wrapfig}
\usepackage{soul}
\usepackage[dvipsnames]{xcolor}
\usepackage{subcaption}
\usepackage{amssymb}
\usepackage{amsmath}
\usepackage{amsthm}
\usepackage{algorithm}
\usepackage{algorithmic}
\usepackage{multirow}
\usepackage{chngpage}

\title{Loss Surfaces, Mode Connectivity, \\ and Fast Ensembling of DNNs}

\author{
	\parbox{\linewidth}{
		\centering
		Timur Garipov\thanks{\text{ }\text{ }Equal contribution.}\, $^{1, 2}$\quad
		Pavel Izmailov$^{*3}$\quad
		Dmitrii Podoprikhin$^{*4}$ \\
		Dmitry Vetrov$^5$\quad
		Andrew Gordon Wilson$^3$
	}\\
	~\\
	\parbox{\linewidth}{
		\centering
		$^1$Samsung AI Center in Moscow,
		$^2$Skolkovo Institute of Science and Technology,\\
		$^3$Cornell University, \\
		$^4$Samsung-HSE Laboratory, National Research University Higher School of Economics,\\
		$^5$National Research University Higher School of Economics
	}
}

\begin{document}

\maketitle

\setcounter{footnote}{0}
\begin{abstract}
The loss functions of deep neural networks are complex and their geometric properties
are not well understood.  We show that the optima of these complex loss functions
are in fact connected by simple curves 
over which training and test accuracy are nearly constant.  We introduce a training procedure to discover these 
high-accuracy pathways between modes.  Inspired by this new geometric insight, 
we also propose a new ensembling method entitled Fast Geometric Ensembling (FGE).
Using FGE we can train 
high-performing ensembles in the time required to train a single model.  We achieve
improved performance compared to 
the recent state-of-the-art Snapshot Ensembles, on  
CIFAR-10, CIFAR-100, and ImageNet. 
\end{abstract}

\section{Introduction}

The loss surfaces of deep neural networks (DNNs) are highly non-convex and can depend on 
millions of parameters.  The geometric properties of these loss surfaces are not well understood.  Even for simple networks, the number of local optima and saddle points is large and can grow exponentially in the number of 
parameters \citep{auer1996exponentially, choromanska2015, dauphin2014identifying}.  Moreover, the loss is
high along a line segment connecting two optima  \citep[e.g.,][]{Goodfellow2015, keskar2017large}.
These two observations suggest that the local optima are isolated.  

\begin{figure*}[!t]
	\centering
	\begin{subfigure}{0.32\textwidth}
		\includegraphics[width=\textwidth]{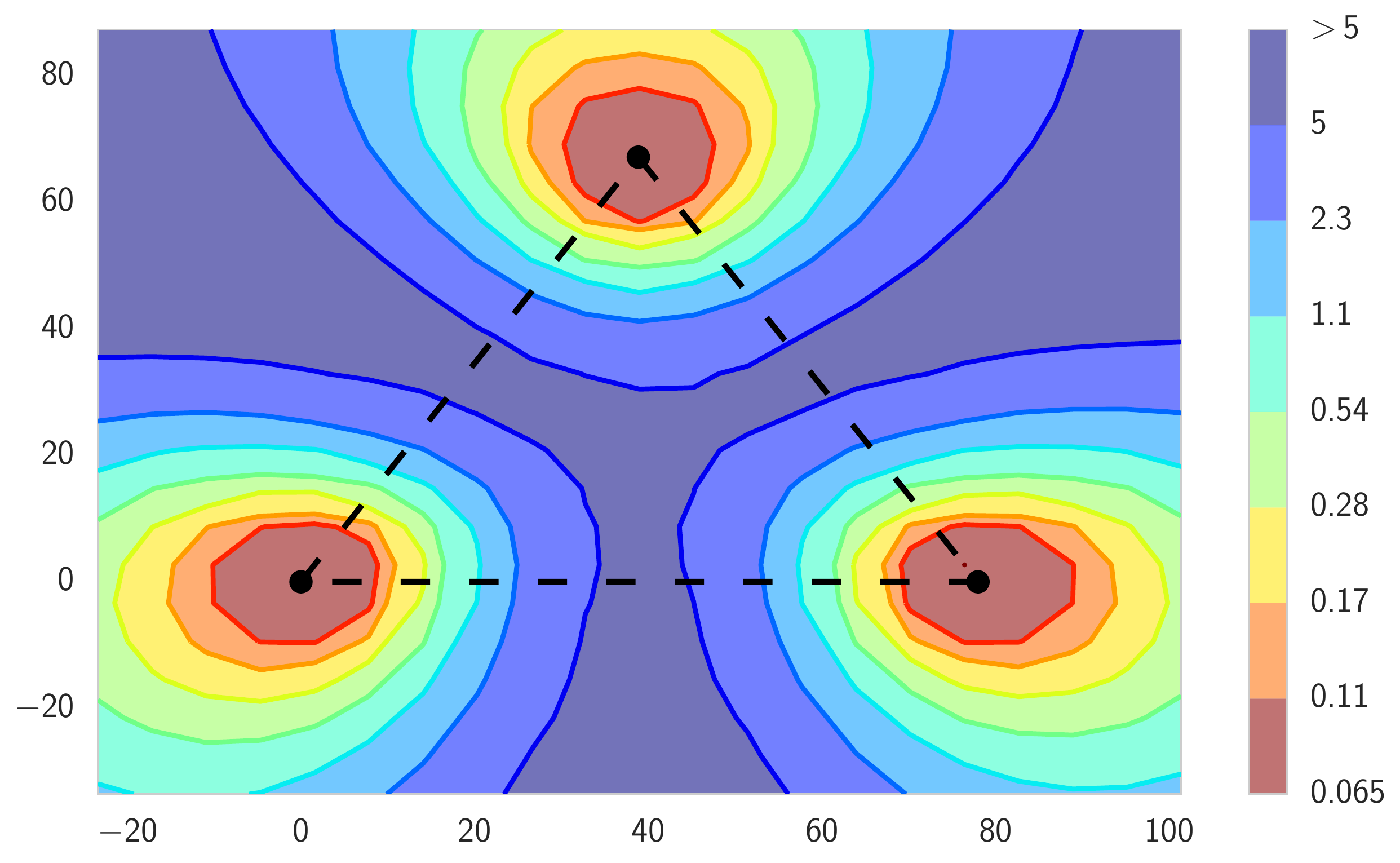}
	\end{subfigure}
	~
	\begin{subfigure}{0.32\textwidth}
		\includegraphics[width=\textwidth]{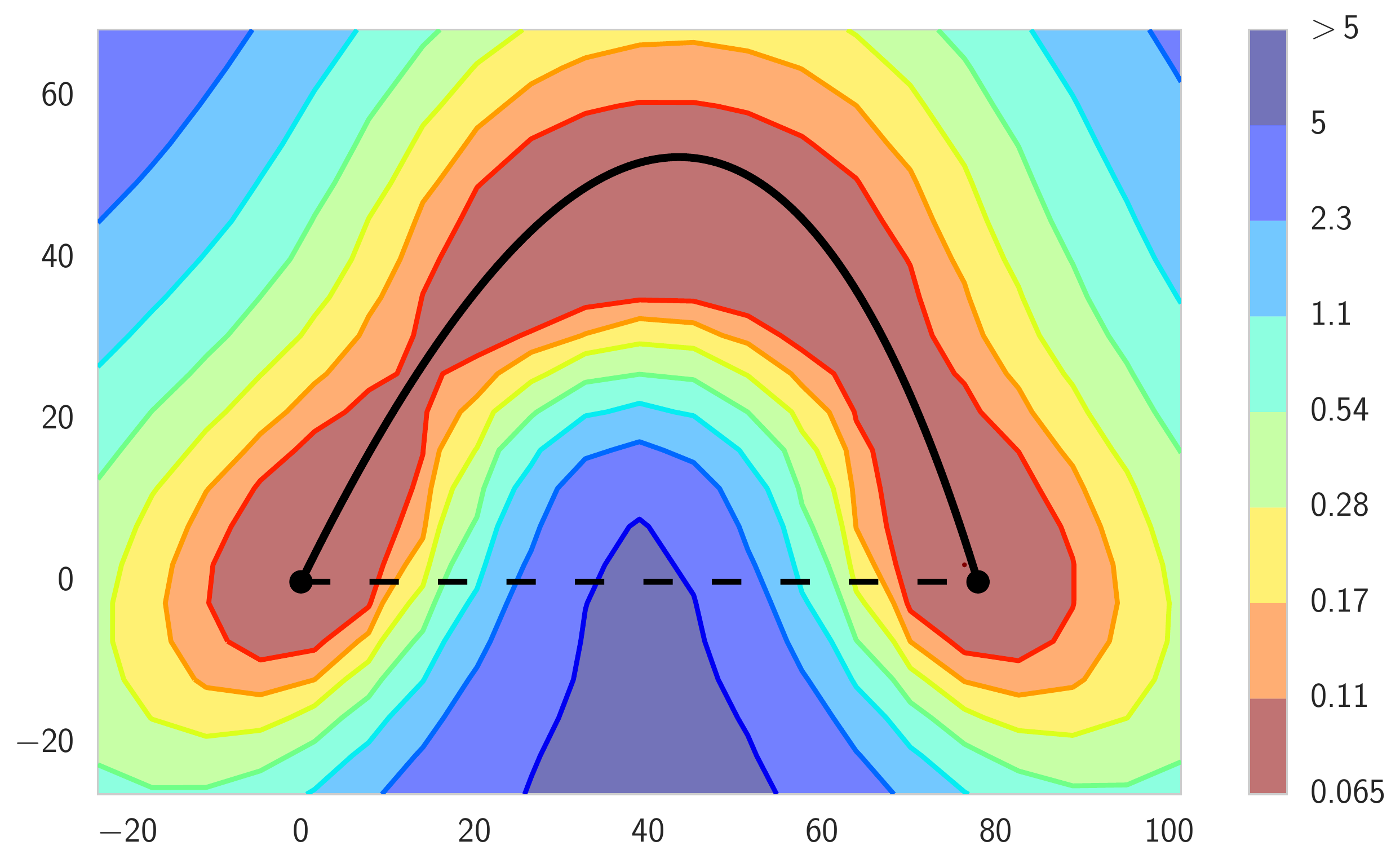}
	\end{subfigure}
	~
	\begin{subfigure}{0.32\textwidth}
		\includegraphics[width=\textwidth]{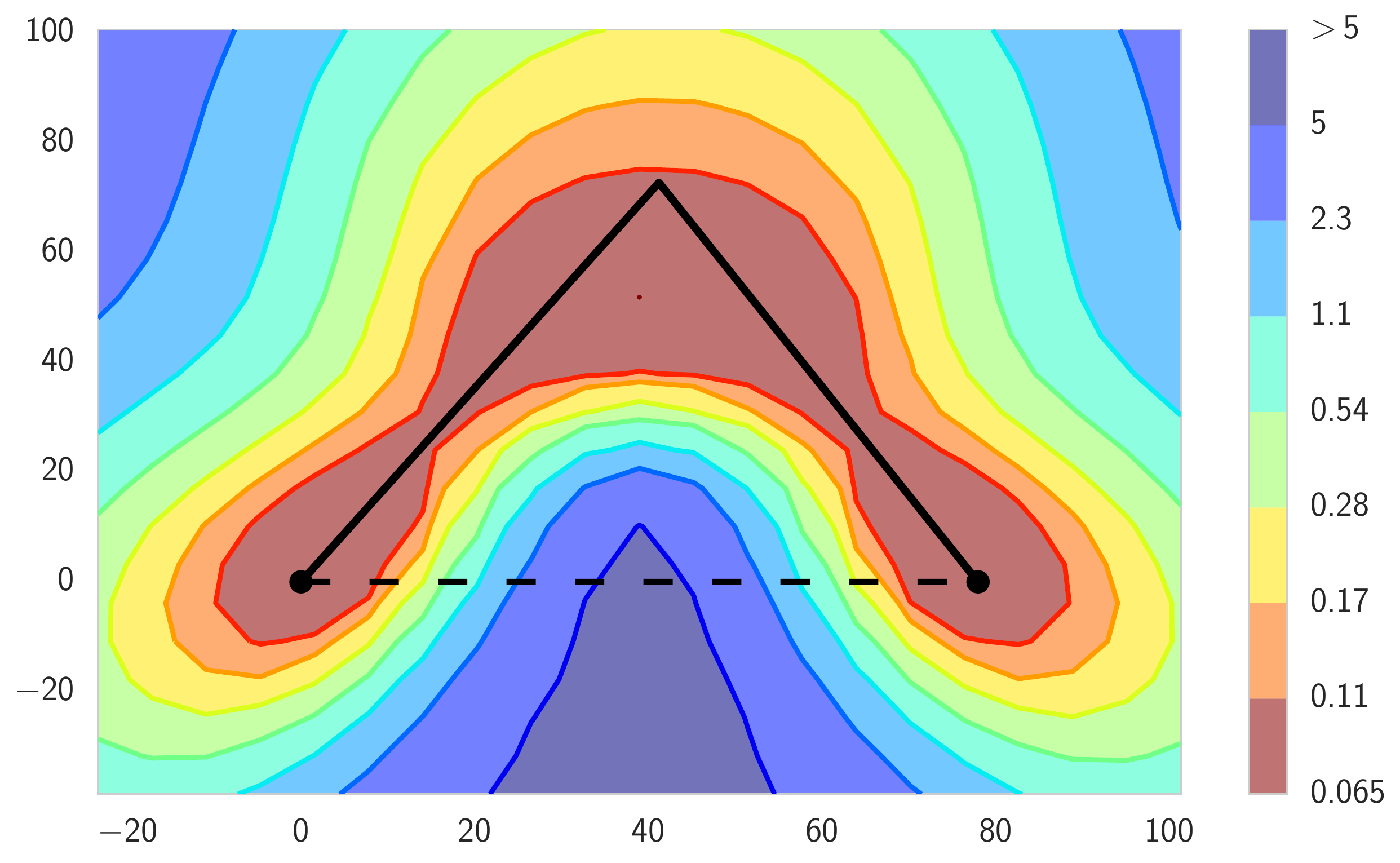}
	\end{subfigure}
	\caption{
		The $\ell_2$-regularized cross-entropy train loss surface of a ResNet-164
		on CIFAR-100, as a function of network weights in a two-dimensional subspace. 
		In each panel, the horizontal axis is fixed and is attached to the optima of two 
		independently trained networks.  The vertical axis changes between panels as 
		we change planes (defined in the main text). \textbf{Left:} Three optima for independently trained networks.
		\textbf{Middle} and \textbf{Right}: A quadratic Bezier curve, and a polygonal
		chain with one bend, connecting the lower two optima on the left panel along
		a path of near-constant loss.  Notice that in each panel a direct 
		linear path between each mode would incur high loss.}
	\label{fig:intro}
\end{figure*}

In this paper, we provide a new training procedure which can in fact find paths of near-constant accuracy 
between the modes of large deep neural networks. Furthermore, we 
show that for a wide range of architectures we can find these paths in the form of a simple polygonal chain of two 
line segments.  Consider, for example, Figure~\ref{fig:intro}, which illustrates the ResNet-164 $\ell_2$-regularized
cross-entropy train loss on CIFAR-100, through three different planes. 
We form each two dimensional plane by all affine combinations of three weight
vectors.\footnote{Suppose we have three weight vectors $w_1, w_2, w_3$. 
We set $u = (w_2 - w_1)$, $v = (w_3 - w_1) -\langle w_3 -~w_1, w_2 - ~w_1 \rangle /  \|w_2 - w_1\|^2 \cdot (w_2 - w_1)$.
Then the normalized vectors $\hat u = u / \|u\|$, $\hat v = v / \|v\|$ form
an orthonormal basis in the plane containing $w_1, w_2, w_3$. To visualize
the loss in this plane, we define a Cartesian grid in the basis $\hat u, \hat v$
and evaluate the networks corresponding to each of the points in the grid. A point
$P$ with coordinates $(x, y)$ in the plane would then be given by 
$P = w_1 + x \cdot \hat u + y \cdot \hat v$.}

The left panel shows
a plane defined by three independently trained networks.  In this plane, all 
optima are isolated, which corresponds to the standard intuition.  However, the middle and 
right panels show two different paths of near-constant loss between the modes in weight space, 
discovered by our proposed training procedure.  
The endpoints of these paths are the two independently trained DNNs corresponding
to the two lower modes on the left panel.

We believe that this geometric discovery has major implications for research into 
multilayer networks, including (1) improving the efficiency, reliability, and accuracy of training, 
(2) creating better ensembles, and (3) deriving more effective posterior approximation 
families in Bayesian deep learning. Indeed, in this paper we are inspired by this geometric
insight to propose a new ensembling procedure that can efficiently discover multiple 
high-performing but diverse deep neural networks.

In particular, our contributions include:
\begin{itemize}
	\item The discovery that the local optima for modern deep neural networks are connected by very simple curves, such as a polygonal chain with only one bend.
	\item A new method that finds such paths between two local optima, 
	such that the train loss and
	test error remain low along these paths.
	\item Using the proposed method we demonstrate that such mode connectivity
	holds for a wide range of modern deep neural networks, on key benchmarks such 
	as CIFAR-100.  We show that these paths correspond to meaningfully different representations that can be efficiently 
	ensembled for increased accuracy. 
		\item Inspired by these observations, we propose Fast Geometric Ensembling (FGE),
	which outperforms the recent state-of-the-art Snapshot Ensembles \citep{huang2017}, on CIFAR-10 and
	CIFAR-100, using powerful deep neural networks 
	such as VGG-16, Wide ResNet-28-10, and ResNet-164. On ImageNet we achieve
	$0.56\%$ top-$1$ error-rate improvement for a pretrained 
  ResNet-50 model by running FGE for only $5$ epochs. 
  \item We release the code for reproducing the results in this paper at \\
\url{https://github.com/timgaripov/dnn-mode-connectivity}
\end{itemize}

The rest of the paper is organized as follows. Section \ref{sec:rel_work} 
discusses existing literature on DNN loss geometry and ensembling techniques.
Section \ref{sec:curves} introduces the proposed method to find 
the curves with low train loss and test error between local optima, which 
we investigate empirically in Section \ref{sec:experiments}.  Section
\ref{sec:fast_ensembling} then introduces our proposed ensembling technique,
FGE, which we empirically compare to the alternatives in Section \ref{sec:fast_ensembling_exp}.
Finally, in Section \ref{sec:future_work} we discuss connections
to other fields and directions for future work.

Note that we interleave two sections where we make methodological 
proposals (Sections \ref{sec:curves}, \ref{sec:fast_ensembling}), 
with two sections where we perform experiments (Sections \ref{sec:experiments}, \ref{sec:fast_ensembling_exp}).
Our key methodological proposal for ensembling, FGE, is in Section \ref{sec:fast_ensembling}.

\section{Related Work}
\label{sec:rel_work}

Despite the success of deep learning across many application domains, the loss 
surfaces of deep neural networks are not well understood.  These loss surfaces 
are an active area of research, which falls into two distinct categories.

The first category explores the local structure of minima found by SGD and its modifications. 
Researchers typically distinguish sharp and 
wide local minima, which are respectively found by using large and small
mini-batch sizes during training. \citet{hochreiter1997flat} and \citet{keskar2017large}, 
for example, claim that flat minima
lead to strong generalization, while sharp minima deliver poor results 
on the test dataset. However, recently \citet{pmlr-v70-dinh17b} argue that most existing 
notions of flatness cannot directly explain generalization. To better understand the local structure of DNN loss minima, 
\citet{li2017visualizing} proposed a new visualization method for 
the loss surface near the minima found by SGD. 
Applying the method for a variety of different architectures, they
showed that the loss surfaces of modern residual networks are 
seemingly smoother than those of VGG-like models.

The other major category of research
considers global loss structure. One of the main questions in this area is 
how neural networks are able to
overcome poor local optima. \citet{choromanska2015} investigated the link 
between the loss function of a simple fully-connected network  and  the  Hamiltonian
of the spherical spin-glass model. Under strong simplifying assumptions they
showed that the values of the investigated loss function at local optima are within
a well-defined bound. In other research, 
\citet{lee2016gradient} showed that under mild conditions gradient descent 
almost surely
converges to a local minimizer and not a saddle point, starting from a random 
initialization.  

In recent work \citet{freeman2017topology}
theoretically show that local minima of a neural network with one hidden layer
and ReLU activations can be connected with a curve along which the loss is 
upper-bounded by a constant that depends on the number
of parameters of the network and the ``smoothness of the data''.  Their 
theoretical results do not readily generalize to multilayer networks.  
Using a dynamic programming approach they empirically construct a polygonal chain for a CNN on MNIST and an RNN on PTB next word prediction. However, in more difficult settings such as AlexNet on CIFAR-10 their approach struggles to achieve even the modest test accuracy of $80\%$. Moreover, they do not consider ensembling.

By contrast, we propose a much simpler training procedure that can 
find near-constant accuracy polygonal chains with only one bend between 
optima, even on a range of modern state-of-the-art architectures. 
Inspired by properties of the loss function discovered by our procedure, we also 
propose a new state-of-the-art ensembling method that can be trained 
in the time required to train a single DNN, with compelling performance on many
key benchmarks (e.g., 96.4\% accuracy on CIFAR-10).

\citet{xie2013horizontal} proposed a related ensembling approach that
gathers outputs of neural networks from different epochs at the end of training 
to stabilize final predictions.  More recently, \citet{huang2017} proposed 
\emph{snapshot ensembles}, which use a cosine cyclical learning rate
\citep{loshchilov2017} to save ``snapshots'' of the model during training at times
when the learning rate achieves its minimum.  In our experiments, we compare
our geometrically inspired approach to \citet{huang2017}, showing improved 
performance.

\citet{draxler2018} simultaneously and independently discovered the existence of 
curves connecting local optima in DNN loss landscapes.
To find these curves they used a different approach inspired by the 
{\it Nudged Elastic Band} method \citep{jonson1998} from quantum chemistry.

\section{Finding Paths between Modes}
\label{sec:curves}

We describe a new method to minimize the training error along a path that connects 
two points in the space of DNN weights.  Section \ref{sec:connection_procedure} 
introduces this general procedure for arbitrary parametric curves, and Section 
\ref{sec:example_parametrizations} describes polygonal chains and Bezier curves
as two example parametrizations of such curves. 
In the supplementary material, we discuss the 
computational complexity of the proposed approach and how to apply 
batch normalization at test time to points on these curves.  We note that after 
curve finding experiments in Section \ref{sec:experiments}, we make our \emph{key 
methodological proposal for ensembling} in Section \ref{sec:fast_ensembling}.

\subsection{Connection Procedure}
\label{sec:connection_procedure}

Let $\hat w_1$ and $\hat w_2$ in $\mathbb{R}^{|net|}$ be two sets of weights
corresponding to two neural networks independently trained by minimizing any
user-specified loss $\mathcal{L}(w)$, such as the cross-entropy loss.  Here, $|net|$ is 
the number of weights of the DNN. Moreover, let
$\phi_{\theta}: [0, 1] \rightarrow \mathbb{R}^{|net|}$ be
a continuous piecewise smooth parametric curve, with parameters
$\theta$, such that 
$ \phi_{\theta}(0) = \hat w_1,~~ \phi_{\theta}(1) = \hat w_2$.  

To find a path of high accuracy between $\hat w_1$ and $\hat w_2$, we propose to find the parameters
$\theta$ that minimize the expectation over a uniform distribution on 
the curve, $\hat \ell (\theta)$:
\begin{equation}
\label{eq:corrected_loss}
\hat \ell (\theta) = 
\frac{\int {\cal L(\phi_{\theta})} d\phi_{\theta}} {\int d\phi_{\theta}} =  
\frac{\int\limits_{0}^{1} {\cal L}(\phi_{\theta}(t))\|\phi'_{\theta}(t)\| dt} {\int\limits_{0}^{1} \|\phi'_{\theta}(t)\| dt} = \int\limits_0^1 \mathcal{L}(\phi_\theta(t)) q_{\theta}(t) dt = \mathbb{E}_{t \sim q_{\theta}(t)} \Big[ \mathcal{L}(\phi_\theta(t)) \Big],
\end{equation}
where the distribution $q_\theta(t)$ on $t \in [0, 1]$ is defined as:
$
q_\theta(t) = \|\phi_{\theta}'(t) \| \cdot \left(\int\limits_{0}^{1} \|\phi_{\theta}'(t)\| dt\right)^{-1}.
$
The numerator of \eqref{eq:corrected_loss} is the line integral of the loss $\mathcal{L}$ on the curve, and the denominator $\int_0^1 \|\phi'_{\theta}(t)\| dt$ is the normalizing constant of the uniform distribution on the curve defined by $\phi_{\theta}(\cdot)$.
Stochastic gradients of $\hat \ell (\theta)$ in 
Eq.~\eqref{eq:corrected_loss} are
generally intractable since $q_\theta(t)$ depends on $\theta$. 
Therefore we also propose a more computationally tractable loss
\begin{equation}
\label{eq:curve_loss}
\ell(\theta) = \int_{0}^{1} \mathcal{L}(\phi_{\theta}(t)) dt =
\mathbb{E}_{t \sim U(0, 1)} \mathcal{L}(\phi_{\theta}(t)),
\end{equation}
where $U(0, 1)$ is the uniform distribution on $[0, 1]$. The difference
between \eqref{eq:corrected_loss} and \eqref{eq:curve_loss} is that the
latter is an expectation of the loss $\mathcal{L}(\phi_\theta(t))$ with
respect to a uniform distribution on $t \in [0, 1]$, while \eqref{eq:corrected_loss}
is an expectation with respect to a uniform distribution on the curve.
The two losses coincide, for example, when $\phi_\theta(\cdot)$ defines a polygonal
chain with two line segments of equal length and the parametrization of 
each of the two segments is linear in $t$. 

To minimize \eqref{eq:curve_loss}, at each iteration we 
sample  $\tilde t$ from the uniform distribution $U(0,1)$ and make a gradient step for 
$\theta$  with respect to the loss $\mathcal{L}(\phi_{\theta}(\tilde t))$. This way we obtain unbiased estimates of the gradients of $\ell(\theta)$, as
\[
 \nabla_\theta \mathcal{L}(\phi_{\theta}(\tilde t)) \simeq  \mathbb{E}_{t \sim U(0, 1)} \nabla_\theta \mathcal{L}(\phi_{\theta}(t)) = 
\nabla_\theta \mathbb{E}_{t \sim U(0, 1)} \mathcal{L}(\phi_{\theta}(t)) = 
\nabla_\theta \ell(\theta).
\]
We repeat these updates until convergence.

\subsection{Example Parametrizations}
\label{sec:example_parametrizations}
\paragraph{Polygonal chain}
\label{sec:polychain}
The simplest parametric curve we consider is the polygonal chain 
(see Figure~\ref{fig:intro}, right). The trained networks 
$\hat w_1$ and $\hat w_2$ serve as the endpoints of the chain and the bends of the chain are
the parameters $\theta$ of the curve parametrization. Consider the simplest case of a chain with 
one bend $\theta$. Then
\[
\phi_{\theta}(t) = 
\left \{
\begin{array}{ll}
2 \left( t \theta + (0.5 - t) \hat w_1\right), & \hspace{-0.2cm} 0 \le t \le 0.5 \\
2 \left((t - 0.5)  \hat w_2 + (1 - t) \theta \right) , & \hspace{-0.2cm} 0.5 \le t \le 1. 
\end{array}
\right .
\]

\paragraph{Bezier curve}
\label{sec:bezier}
A Bezier curve (see Figure~\ref{fig:intro}, middle)
provides a convenient parametrization of smooth paths with given endpoints. 
A quadratic Bezier curve $\phi_{\theta}(t)$ with
endpoints $\hat w_1$ and $\hat w_2$ is given by
\[
\phi_{\theta}(t) = 
(1 -t)^2 \hat w_1 + 2t(1-t) \theta + t^2 \hat w_2,~~ 0 \le t \le 1.
\]

These formulas naturally generalize for $n$ bends $\theta~=~\{w_1, w_2, \ldots, w_n\}$ 
(see supplement).

\begin{figure}[!h]
	\centering
	~
	\begin{subfigure}{0.31\textwidth}
		\includegraphics[width=\textwidth]{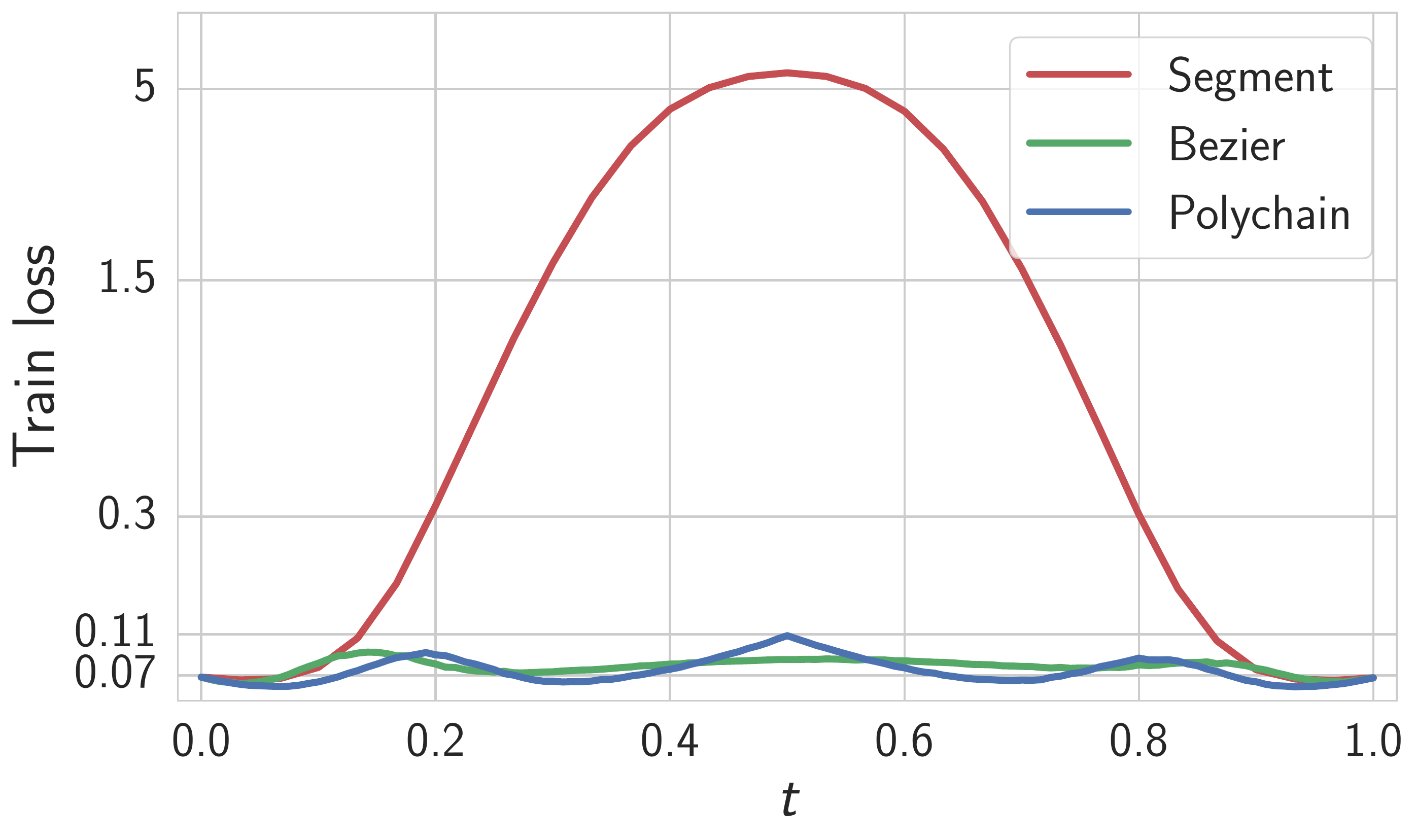}
	\end{subfigure}
	~
	\begin{subfigure}{0.31\textwidth}
		\includegraphics[width=\textwidth]{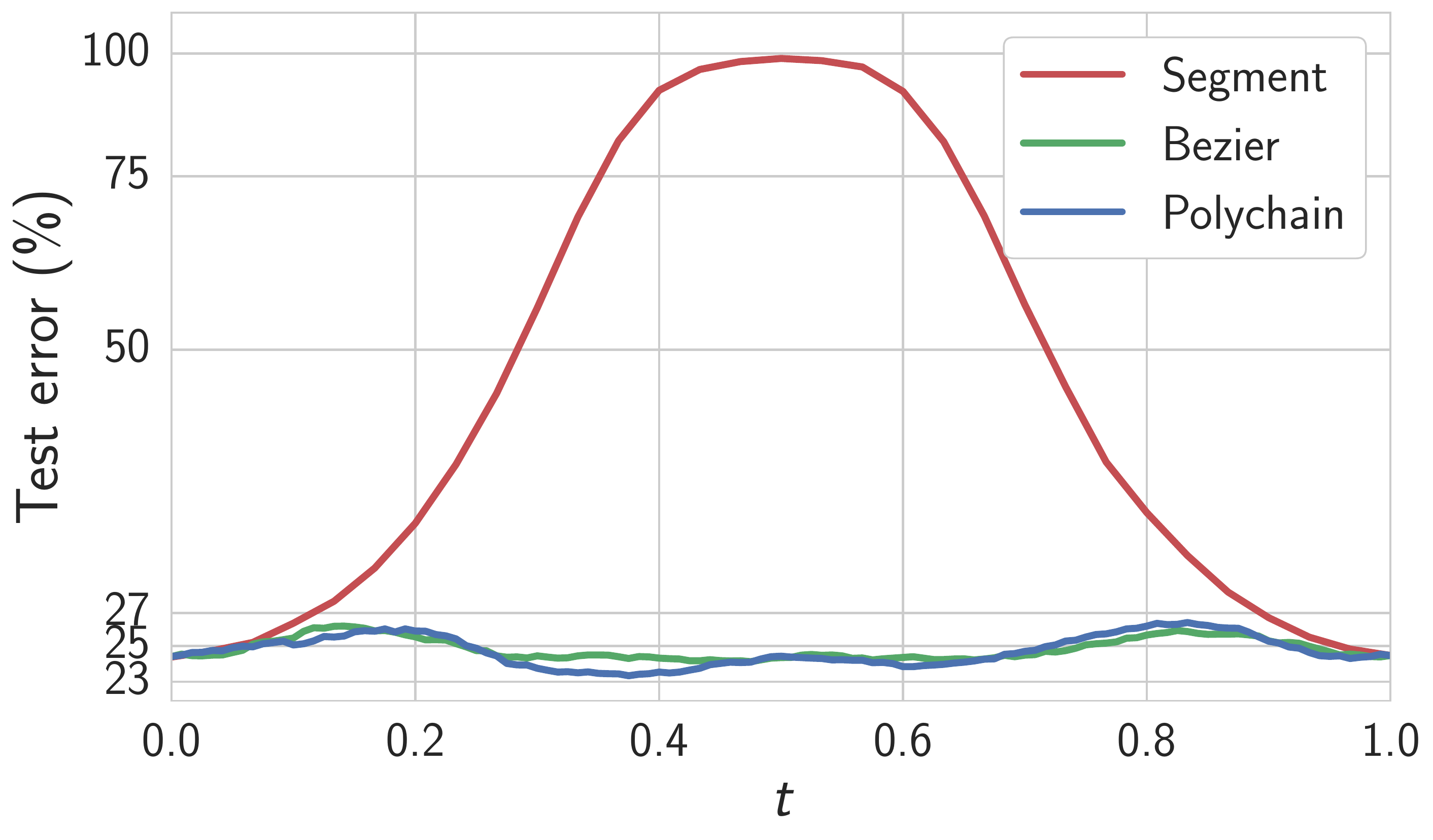}
	\end{subfigure}
	~
	\begin{subfigure}{0.31\textwidth}
		\centering
		\includegraphics[width=\textwidth]{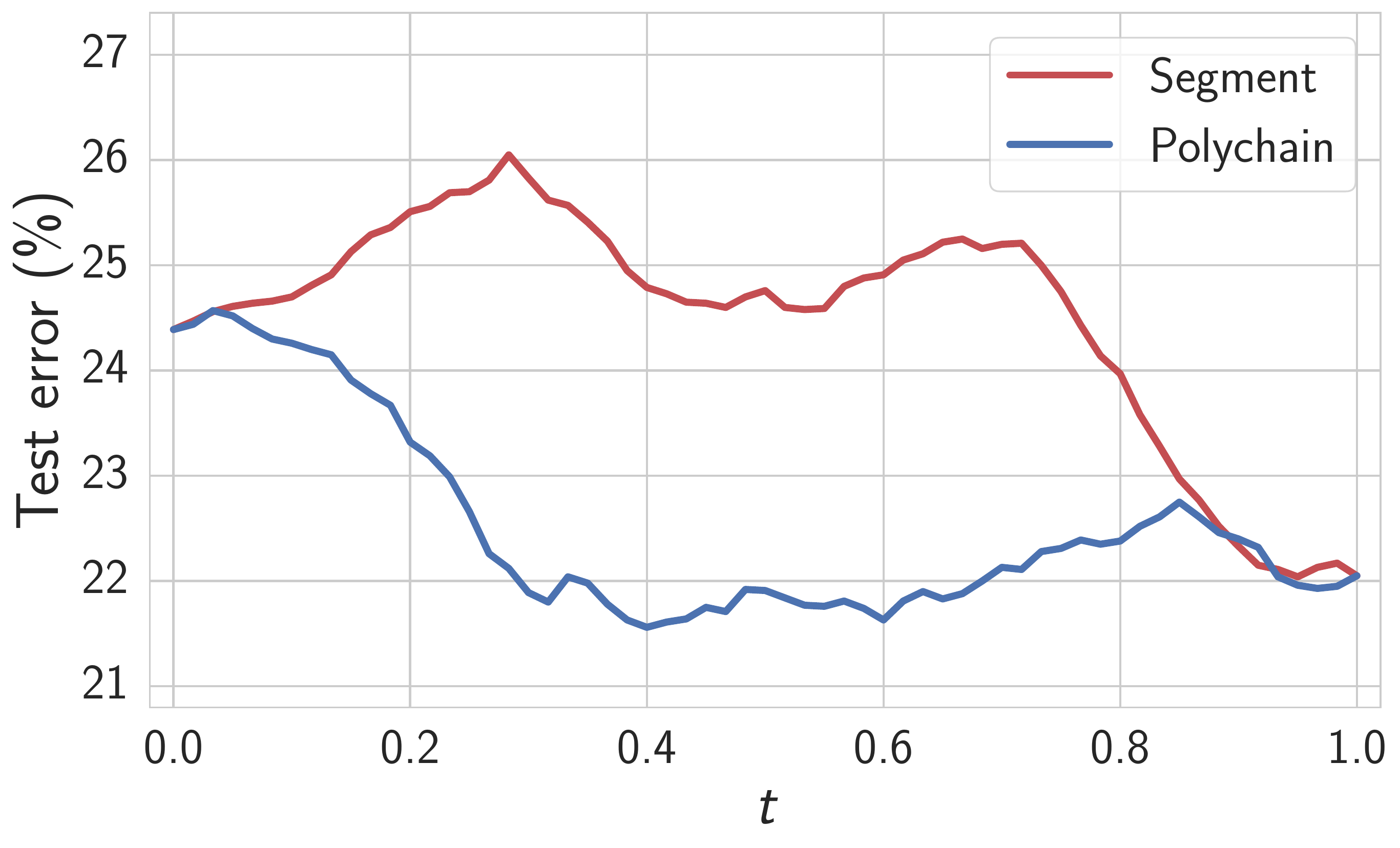}	
	\end{subfigure}
	\caption{
    The $\ell_2$-regularized cross-entropy train loss {\bfseries (left)} 
    and  test error {\bfseries (middle)} as a 
    function of the point on the curves $\phi_{\theta}(t)$ found
    by the proposed method (ResNet-164 on CIFAR-100).
		{\bfseries Right:} Error of the two-network ensemble consisting of the endpoint $\phi_\theta(0)$ of 
		the curve and the point $\phi_\theta(t)$ on the curve (CIFAR-100, ResNet-164).
		``Segment'' is a line segment connecting two modes found by SGD. 
		``Polychain'' is a polygonal chain connecting the same endpoints. 
	}		
	\label{fig:curves_eval}
\end{figure}

\section{Curve Finding Experiments}
\label{sec:experiments}

We show that the proposed training procedure in Section~\ref{sec:curves}
does indeed find high accuracy paths connecting different modes, across 
a range of architectures and datasets. Moreover,
we further investigate the properties of these curves, showing that they 
correspond to meaningfully different representations that can be ensembled
for improved accuracy.  We use these insights to propose an improved ensembling
procedure in Section \ref{sec:fast_ensembling}, which we empirically validate in 
Section~\ref{sec:fast_ensembling_exp}.

In particular, we test VGG-$16$ \citep{simonyan2014very}, a
$28$-layer Wide ResNet with widening factor $10$ \citep{zagoruyko2016wide}
and a $158$-layer ResNet \citep{he2016deep} on CIFAR-10, and VGG-$16$, $164$-layer ResNet-bottleneck \citep{he2016deep} on CIFAR-100. For CIFAR-10
and CIFAR-100 we use the same standard data augmentation as \citet{huang2017}. 
We provide additional
results, including detailed experiments for fully connected and recurrent networks, in the 
supplement.  

For each model and dataset we train two networks with different random 
initializations to find two modes. Then we use the proposed algorithm of Section \ref{sec:curves} 
to find a path connecting these two modes in the weight space with a quadratic 
Bezier curve and a polygonal chain with one bend. We also connect the two modes
with a line segment for comparison. In all experiments we optimize the loss 
\eqref{eq:curve_loss}, as for Bezier curves the gradient of loss 
\eqref{eq:corrected_loss} is intractable, and for polygonal chains we found 
loss \eqref{eq:curve_loss} to be more stable. 

Figures \ref{fig:intro} and \ref{fig:curves_eval} show the results of the proposed mode connecting 
procedure for ResNet-164 on CIFAR-100.  Here \emph{loss} refers to $\ell_2$-regularized
cross-entropy loss. 
For both the Bezier curve and polygonal chain, train loss (Figure~\ref{fig:curves_eval}, left) and
test error (Figure~\ref{fig:curves_eval}, middle) are indeed nearly constant. In addition, we provide plots
of train error and test loss in the supplementary material.
In the supplement, we also include a comprehensive table summarizing all path finding experiments on 
CIFAR-10 and CIFAR-100 for VGGs, ResNets and Wide ResNets, as well as fully connected networks and 
recurrent neural networks, which follow the same general trends. In the supplementary
material we also show that the connecting 
curves can be found consistently as we vary the number of parameters 
in the network, although the ratio of the arclength for the curves to the length
of the line segment connecting the same endpoints decreases with increasing parametrization.
In the supplement, we also measure the losses
\eqref{eq:corrected_loss}  and \eqref{eq:curve_loss} for all the curves we 
constructed, and find that the values of the two losses are very close,
suggesting that the loss \eqref{eq:curve_loss} is a good practical approximation 
to the loss \eqref{eq:corrected_loss}.

The constant-error curves connecting two given networks 
discovered by the proposed method are not unique. We trained two different 
polygonal chains with the same endpoints and
different random seeds using VGG-16 on CIFAR-10. We then measured the
Euclidean distance between the turning points of these curves. 
For VGG-16 on CIFAR-10 this distance is equal to $29.6$ and the distance between 
the endpoints is $50$, showing that the curves are not unique. In this instance, we expect
the distance between turning points to be less than the distance between endpoints,
since the locations of the turning points were initialized to the same value (the center
of the line segment connecting the endpoints).

Although high accuracy connecting curves can often be very simple, such as a polygonal chain
with only one bend, we note that line segments \emph{directly} connecting two modes generally incur high 
error. For VGG-16 on CIFAR-10 the test error goes up to $90\%$ in the center of the
segment. For ResNet-158 and Wide ResNet-28-10 the worst errors along
direct line segments are still high, but relatively less, at $80\%$ and $66\%$, respectively.
This finding suggests that the loss surfaces of state-of-the-art residual networks are 
indeed more regular than those of classical models like VGG, in accordance with
the observations in \citet{li2017visualizing}.

In this paper we focus on connecting pairs of networks trained using the same
hyper-parameters, but from different random initializations.
Building upon our work, \citet{gotmare2018using} have 
recently shown that our mode connectivity approach applies to pairs of networks
trained with different batch sizes, optimizers, data augmentation strategies,
weight decays and learning rate schemes.

To motivate the ensembling procedure proposed in the next
section, we now examine how far we 
need to move along a connecting curve to find a point that produces 
substantially different, but still useful, predictions.
Let $\hat w_1$ and $\hat w_2$ be two 
distinct sets of weights corresponding to optima obtained
by independently training a DNN two times.
We have shown that there
exists a path connecting $\hat w_1$ and $\hat w_2$ with 
high test accuracy. Let $\phi_{\theta}(t)$, $t \in [0, 1]$ parametrize
this path with $\phi_{\theta}(0) = \hat w_1$, $\phi_{\theta}(1) = \hat w_2$.
We investigate the performance of an ensemble of two networks: the
endpoint $\phi_\theta(0)$ of the curve and a point $\phi_\theta(t)$ on the curve corresponding 
to $t \in [0, 1]$. Figure \ref{fig:curves_eval} (right) shows the test error
of this ensemble as a function of $t$, for a ResNet-164 on CIFAR-100.  
The test error starts decreasing at $t \approx 0.1$
and for $t \ge 0.4$ the error of an ensemble is already as low as 
the error of an ensemble of the two independently trained networks used as
the endpoints of the curve. Thus even by moving away from the 
endpoint by a relatively small distance along the curve we can find a network that
produces meaningfully different predictions from the network at the endpoint.
This result also demonstrates that these curves do not exist only due to degenerate
parametrizations of the network (such as rescaling on either side of a ReLU); instead, points 
along the curve correspond to meaningfully different representations of the data that can be 
ensembled for improved performance. In the supplementary
material we show how to create trivially connecting curves that do not have this property.

\section{Fast Geometric Ensembling}
\label{sec:fast_ensembling}

In this section, we introduce a practical ensembling procedure, Fast Geometric Ensembling (FGE), 
motivated by our observations about mode connectivity.

\begin{figure}[!h]
	\centering
	\begin{subfigure}{0.47\textwidth}
		\centering
		\includegraphics[width=\textwidth]{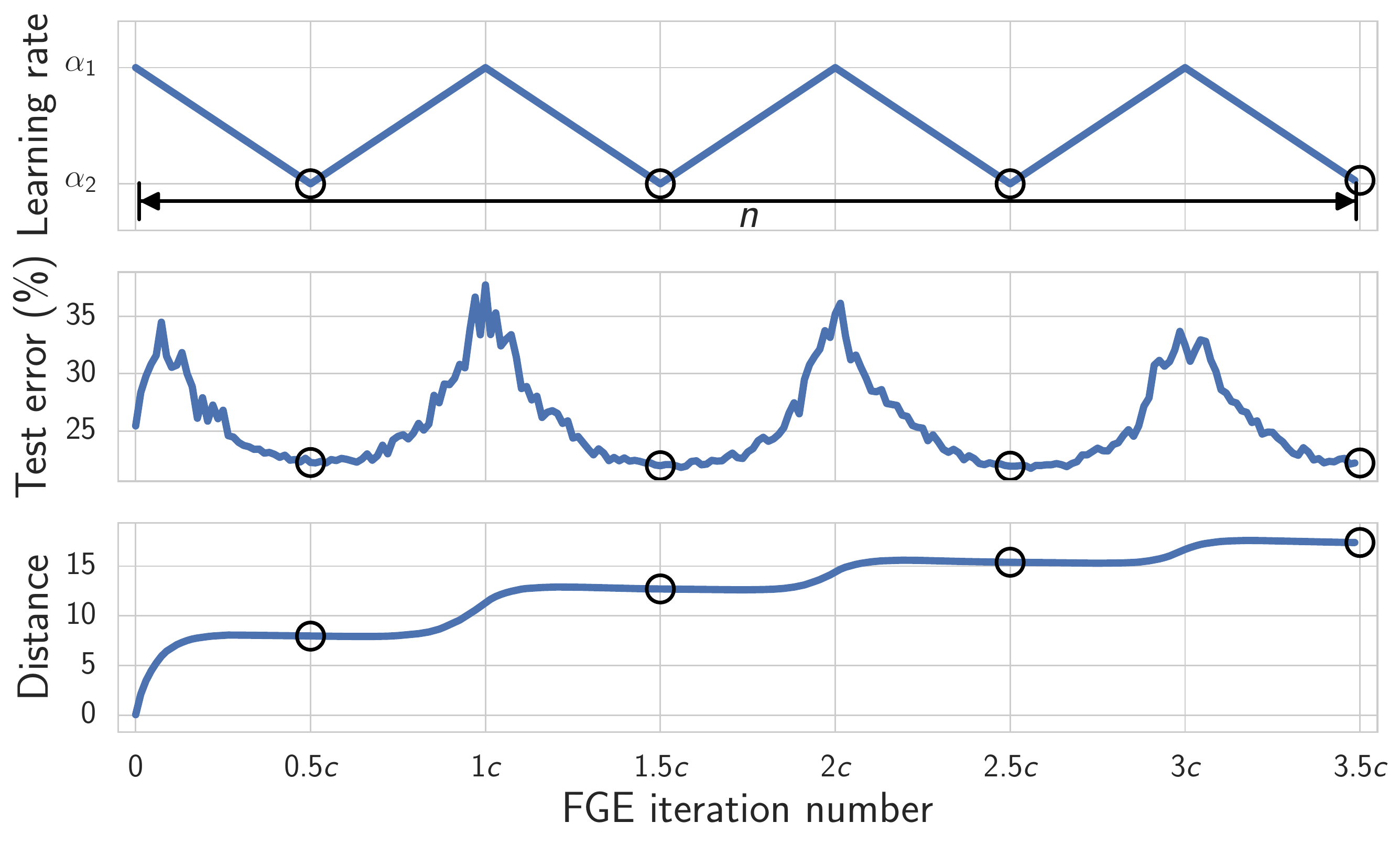}
	\end{subfigure}
	\qquad
	\begin{subfigure}{0.47\textwidth}
		\centering
		\includegraphics[width=\textwidth]{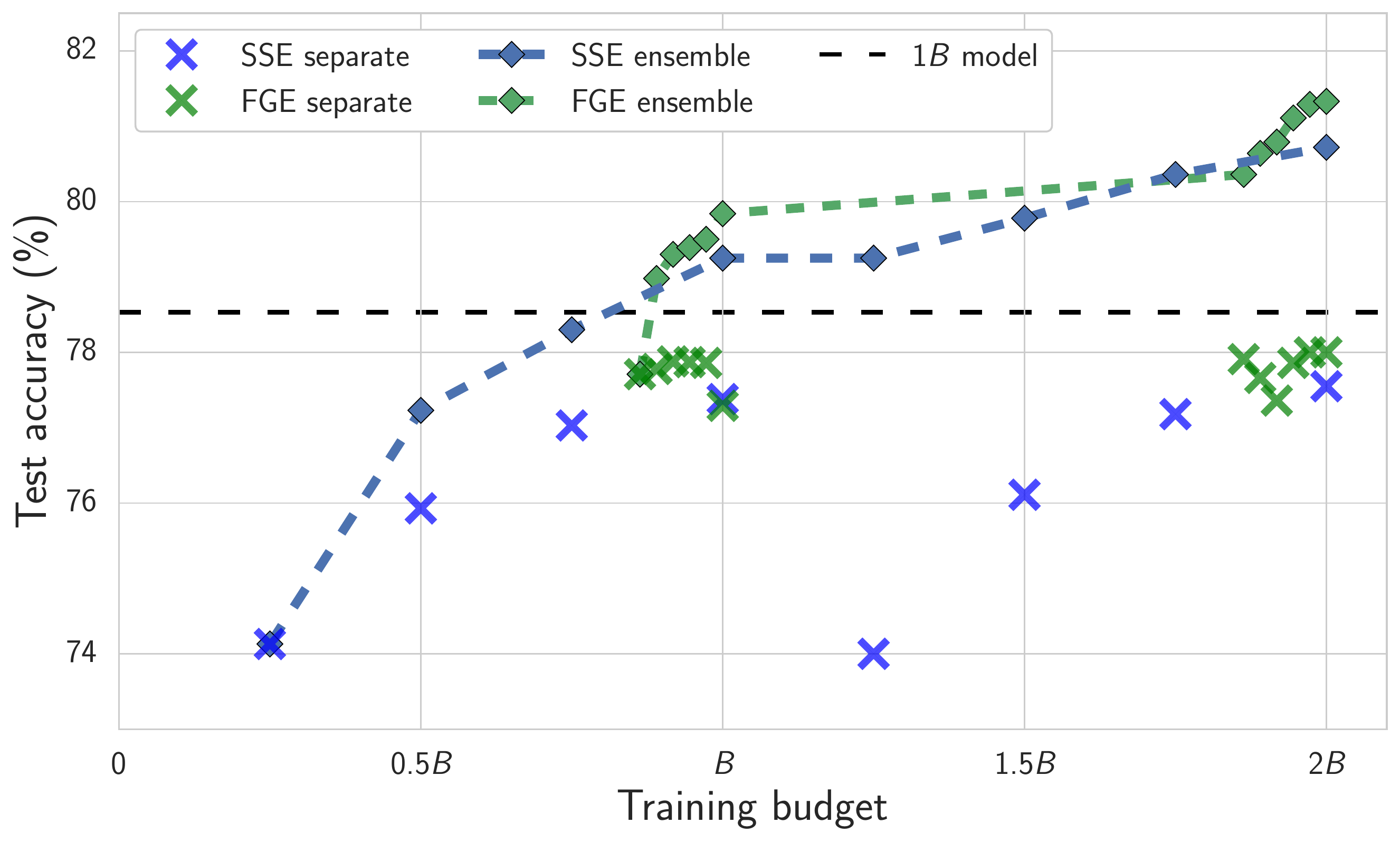}  	
	\end{subfigure}
	\caption{		
		{\bfseries Left:} Plot of the learning rate (\textbf{Top}), 
		test error (\textbf{Middle}) and distance from the
		initial value $\hat w$ (\textbf{Bottom}) as a function of iteration
		for FGE with Preactivation-ResNet-164 on CIFAR-100.
		Circles indicate the times when we save models for ensembling.		
		{\bfseries Right:} Ensemble performance of FGE 
		and SSE (Snapshot Ensembles) as a function of training
		time, using ResNet-164 on CIFAR-100 ($B = 150$ epochs). Crosses represent the performance of separate ``snapshot''
		models, and diamonds show the performance of the ensembles
		constructed of all models available by the given time.
	}
	\label{fig:FGE-combined}    		
\end{figure}

In the previous section, we considered ensembling along mode connecting
curves. Suppose now we instead only have \textbf{one} set of weights $\hat w$ 
corresponding to a mode of the loss. We cannot explicitly construct a path $\phi_{\theta}(\cdot)$
as before, but we know that multiple paths passing through $\hat w$ exist,
and it is thus possible to move away from $\hat w$ in the weight
space without increasing the loss. Further, we know that we can find
diverse networks providing meaningfully different predictions by making
relatively small steps in the weight space (see Figure \ref{fig:curves_eval}, right).  

Inspired by these observations,
we propose the Fast Geometric Ensembling (FGE) method that aims to find diverse 
networks with 
relatively small steps in the weight space, without leaving a region
that corresponds to low test error. 

While inspired by mode connectivity, 
FGE {\bf does not} rely on explicitly finding a connecting curve, and thus 
does not require pre-trained endpoints, and so can be trained in the time 
required to train a \emph{single} network. 

Let us describe Fast Geometric Ensembling. First, we initialize a copy of the network with weights $w$ set equal to the 
weights of the trained network $\hat w$. Now, to force $w$ to move away from
$\hat w$ without substantially decreasing the prediction accuracy we adopt a 
cyclical learning rate schedule $\alpha(\cdot)$ (see Figure \ref{fig:FGE-combined}, left), with the learning rate at iteration $i = 1, 2, \ldots$ defined as
\[
\begin{gathered}
\alpha(i) = \left\{ 
\begin{array}{ll}
(1 - 2 t(i)) \alpha_1 + 2 t(i) \alpha_2 & 0 < t(i) \le \frac{1}{2} \\
(2 - 2 t(i)) \alpha_2 + (2 t(i) - 1) \alpha_1 & \frac{1}{2} < t(i) \le 1 \\
\end{array} 
\right.,
\end{gathered}
\] 
where $t(i) = \frac{1}{c}(\bmod(i - 1, c) + 1)$, the learning rates are
$\alpha_1 > \alpha_2$, and the number of iterations
in one cycle is given by even number $c$. Here by \textbf{iteration} we
mean processing one mini-batch of data.  
We can train the
network $w$ using the standard $\ell_2$-regularized cross-entropy loss function
(or any other loss that can be used for DNN training) with 
the proposed learning rate schedule for $n$ iterations.
In the middle of each learning rate cycle when the learning rate reaches
its minimum value $\alpha(i) = \alpha_2$ (which corresponds to $\bmod(i - 1,c) + 1=c/2, t(i)=\frac{1}{2}$)
we collect the checkpoints of weights $w$. When the training is finished we 
ensemble the collected models. An outline of the algorithm is provided in 
the supplement.

Figure \ref{fig:FGE-combined} (left) illustrates the adopted learning rate schedule.
During the periods when the learning rate is large (close to $\alpha_1$), $w$ is 
exploring the weight space doing larger steps but sacrificing the test
error. When the learning rate is small (close to $\alpha_2$), $w$ is in
the exploitation phase in which the steps become smaller and the test error
goes down. The cycle length is usually about $2$ to $4$ epochs,
so that the method efficiently balances exploration and exploitation
with relatively-small steps in the weight space that are still sufficient
to gather diverse and meaningful networks for the ensemble.

To find a good initialization $\hat w$ for the proposed procedure, we first
train the network with the standard learning rate schedule (the schedule used 
to train single DNN models) for about $80\%$ of the time required to train
a single model. After this pre-training is finished we initialize FGE 
with $\hat w$ and run the proposed 
fast ensembling algorithm for the remaining computational
budget.
In order to get more diverse samples, one can run the algorithm described above 
several times for a smaller number of iterations
initializing from different checkpoints saved during training of 
$\hat w$, and then ensemble all of the 
models gathered across these runs.

Cyclical learning rates have also recently been considered in \citet{smith2017exploring} and \citet{huang2017}.
Our proposed method is perhaps most closely related to Snapshot Ensembles \citep{huang2017}, but has 
several distinctive features, inspired by our geometric insights.
In particular, Snapshot Ensembles adopt cyclical learning rates with
cycle length on the scale of $20$ to $40$ epochs
from the beginning of the training as they are trying to do large steps in the 
weight space. However, according to our analysis of the curves it is sufficient
to do relatively small steps in the weight space to get diverse networks, so
we only employ cyclical learning rates with a small cycle length on the scale of
$2$ to $4$ epochs in the last stage of the training.  
As illustrated in Figure \ref{fig:FGE-combined} (left), the step sizes made by FGE between saving
two models (that is the euclidean distance between sets of weights of corresponding models in the weight space) are on the scale of $7$ for Preactivation-ResNet-164 on
CIFAR-100. For 
Snapshot Ensembles for the same model the distance between two snapshots 
is on the scale of $40$.
We also use a piecewise
linear cyclical learning rate schedule following \citet{smith2017exploring}
as opposed to the cosine schedule in Snapshot Ensembles. 

\section{Fast Geometric Ensembling Experiments}
\label{sec:fast_ensembling_exp}

\begin{table}[!h]
	\caption{Error rates ($\%$) on CIFAR-100 and CIFAR-10 datasets for different ensembling techniques and training budgets. The best results for each dataset, architecture, and budget are {\bfseries bolded}.
	}
	\label{table:fast_ensembling}
	\centering
	\begin{tabular}{clcccccc}
		\toprule
		& & \multicolumn{3}{c}{CIFAR-100}   & \multicolumn{3}{c}{CIFAR-10}\\
		\cmidrule(lr){3-5} 	\cmidrule(lr){6-8}
		DNN (Budget) & method & $1B$ & $2B$ & $3B$ & $1B$ & $2B$ & $3B$ \\
		\midrule
		\multirow{3}{*}{VGG-16 ($200$)} 
		& Ind & $27.4 \pm 0.1$ & $25.28$ & $24.45$ & $6.75 \pm 0.16$ & $5.89$ & $5.9$ \\ 
		& SSE  & $26.4 \pm 0.1$ & $25.16$ & $24.69$  & $6.57 \pm 0.12$ & $6.19$ & $5.95$ \\  
		& FGE & $\mathbf{25.7 \pm 0.1}$ & $\mathbf{24.11}$ & $\mathbf{23.54}$ & $\mathbf{6.48 \pm 0.09}$ & $\mathbf{5.82}$ & $\mathbf{5.66}$ \\ 
		
		\midrule
		
		\multirow{3}{*}{ResNet-164 ($150$)} 
		& Ind & $21.5 \pm 0.4$ & $19.04$ & $18.59$ & $4.72 \pm 0.1$ & $\mathbf{4.1}$ & $\mathbf{3.77}$ \\ 
		& SSE & $20.9 \pm 0.2$ & $19.28$ & $18.91$ & $4.66 \pm 0.02$ & $4.37$ & $4.3$ \\ 
		& FGE  & $\mathbf{20.2 \pm 0.1}$ & $\mathbf{18.67}$ & $\mathbf{18.21}$ & $\mathbf{4.54 \pm 0.05}$ & $4.21$ & $3.98$\\ 
		
		\midrule
		
		\multirow{3}{*}{WRN-28-10 ($200$)} 
		& Ind & $19.2 \pm 0.2$ & $17.48$ & $17.01$ & $3.82 \pm 0.1$ & $3.4$ & $\mathbf{3.31}$ \\ 
		& SSE & $17.9 \pm 0.2$ & $17.3$ & $16.97$ & $3.73 \pm 0.04$ & $3.54$ & $3.55$ \\ 
		& FGE  & $\mathbf{17.7 \pm 0.2}$ & $\mathbf{16.95}$ & $\mathbf{16.88}$ & $\mathbf{3.65 \pm 0.1}$ & $\mathbf{3.38}$ & $3.52$ \\
		\bottomrule
	\end{tabular}
\end{table}

In this section we compare the proposed Fast Geometric Ensembling (\textbf{FGE}) technique against
ensembles of independently trained networks (\textbf{Ind}), and SnapShot Ensembles (\textbf{SSE}) \citep{huang2017},
a recent state-of-the-art fast ensembling approach.   

For the ensembling experiments we use a $164$-layer Preactivation-ResNet
in addition to the VGG-16 and Wide ResNet-28-10 models.  Links for implementations to 
these models can be found in the supplement. 

We compare the accuracy of each method as a function of computational budget.
For each network architecture and dataset we denote the number of epochs 
required to train
a single model as $B$. 
For a $kB$ budget, we run each of Ind, FGE and SSE   
$k$ times from random initializations
and ensemble the models gathered from the $k$ runs.
In our experiments we set $B = 200$ for VGG-16 and Wide ResNet-28-10 (WRN-28-10) models, and
$B=150$ for ResNet-164, since $150$ epochs is typically sufficient to train this model.  We note
the runtime per epoch for FGE, SSE, and Ind is the same, and so the total computation associated 
with $kB$ budgets is the same for all ensembling approaches.

For Ind, we use an initial learning rate of 0.1 for ResNet and Wide ResNet, and 0.05 for VGG.  
For FGE, with VGG we use cycle length $c = 2$ epochs, and a total of $22$
models in the final ensemble. With ResNet and Wide ResNet we use $c = 4$ epochs, and the total 
number of models in the final ensemble is $12$ for Wide ResNets and $6$ for ResNets. 
For VGG we set the learning rates to $\alpha_1 = 10^{-2}$, $\alpha_2 = 5 \cdot 10^{-4}$; for ResNet
and Wide ResNet models we set $\alpha_1 = 5 \cdot 10^{-2}$, 
$\alpha_2 = 5 \cdot 10^{-4}$. .
For SSE, we followed \citet{huang2017} and varied the initial learning rate
$\alpha_0$ and number of snapshots per run $M$.
We report the best results we achieved, which corresponded to 
$\alpha_0 = 0.1, M = 4$ for ResNet, 
$\alpha_0 = 0.1, M = 5$ for Wide ResNet, and $\alpha_0 = 0.05, M = 5$ for VGG.  The total number of models in the FGE ensemble is constrained by network
choice and computational budget. 
Further experimental details are in the supplement.

Table \ref{table:fast_ensembling} summarizes the results of the experiments. 
In all conducted experiments FGE outperforms SSE, particularly as we increase
the computational budget. The 
performance improvement against Ind is most noticeable for CIFAR-100.  With 
a large number of classes, any two models are less likely to make the same predictions.
Moreover, there will be greater uncertainty over which representation one should use on 
CIFAR-100, since the number of classes is increased tenfold from CIFAR-10, but the number
of training examples is held constant.  Thus
smart ensembling strategies will be especially important on this dataset.
Indeed in all experiments on CIFAR-100, FGE outperformed all other 
methods. On CIFAR-10, FGE consistently improved upon SSE for all budgets and architectures.  
FGE also improved against Ind for all training budgets with VGG, but is more similar in performance
to Ind on CIFAR-10 when using ResNets.

Figure~\ref{fig:FGE-combined} (right) illustrates the results for 
Preactivation-ResNet-164 on CIFAR-100 for one and two training
budgets. The training budget $B$ is $150$ epochs. Snapshot Ensembles use a cyclical learning rate from the 
beginning of the training and they gather the models for the
ensemble throughout training.
To find a good initialization
we run standard independent 
training for the first $125$ epochs 
before applying FGE.
In this case, the whole ensemble is gathered over the following $22$ epochs ($126$-$147$) to fit in the budget of each of the two runs.
During these $22$ epochs FGE is able to gather diverse enough networks to
outperform Snapshot Ensembles both for $1 B$ and $2 B$ budgets.

Diversity of predictions of the individual networks is crucial for the 
ensembling performance \citep[e.g.,][]{{lee2016stochastic}}.
We note that the diversity of the networks averaged by FGE is lower than that of
completely independently trained networks. Specifically, two independently trained 
ResNet-164 on CIFAR-100 make different predictions on $19.97\%$ of test objects, while 
two networks from the same FGE run make different predictions on $14.57\%$ of test objects. 
Further, performance of
individual networks averaged by FGE is slightly lower than that of fully trained networks (e.g. $78.0\%$ against $78.5\%$
on CIFAR100 for ResNet-164). 
However, for a given computational budget FGE can propose many 
more high-performing networks than independent training, leading to better 
ensembling performance (see Table \ref{table:fast_ensembling}). 

\subsection{ImageNet}

ImageNet ILSVRC-2012 \citep{russakovsky2015imagenet}
is a large-scale dataset containing  $1.2$  million  training  
images  and  $50000$ validation images divided into $1000$ classes.

CIFAR-100 is the primary focus of our ensemble experiments. However,
we also include ImageNet results for the proposed FGE procedure, using a 
ResNet-50 architecture. We used a pretrained model with top-$1$ test error
of $23.87$ to initialize the FGE procedure. We then ran FGE for $5$ epochs
with a cycle length of $2$ epochs and with learning rates $\alpha_1 = 10^{-3}$,
$\alpha_2 = 10^{-5}$. The top-$1$ test error-rate of the final ensemble was 
$23.31$. Thus, in just $5$ epochs we could improve the accuracy of the 
model by $0.56$ using FGE. The final ensemble contains $4$ models (including the pretrained one).
Despite the harder setting of only $5$ epochs to
construct an ensemble, FGE performs comparably to the best result reported by 
\citet{huang2017} on ImageNet, $23.33$ error, which was also achieved using a ResNet-50.

\section{Discussion and Future Work}
\label{sec:future_work}

We have shown that the optima of deep neural networks are connected 
by simple pathways, such as a polygonal chain with a single bend, 
with near constant accuracy.  We introduced a training
procedure to find these pathways, with a user-specific curve of choice.
We were inspired by these insights to propose a practical new ensembling
approach, Fast Geometric Ensembling, which achieves state-of-the-art 
results on CIFAR-10, CIFAR-100, and ImageNet.  

There are so many exciting future directions for this research. At a high 
level we have shown that even though the loss surfaces of deep neural 
networks are very complex, there is relatively simple structure connecting
different optima.  Indeed, we can now move towards thinking about 
valleys of low loss, rather than isolated modes.  

These valleys could 
inspire new directions for approximate Bayesian inference, such as 
stochastic MCMC approaches which could now jump along these bridges
between modes, rather than getting stuck exploring a single mode.  One 
could similarly derive new proposal distributions for variational inference, 
exploiting the flatness of these pathways.  These geometric insights could also
be used to accelerate the convergence, stability and 
accuracy of optimization procedures like SGD, by helping us understand the 
trajectories along which the optimizer moves, and making it possible to develop procedures 
which can now search in more structured spaces of high accuracy.   One could
also use these paths to construct methods which are more robust to adversarial
attacks, by using an arbitrary collection of diverse models described by a high 
accuracy curve, returning the predictions of a different model for each query from
an adversary. We can also use this new property to create better visualizations 
of DNN loss surfaces.  Indeed, using the proposed training procedure, we were
able to produce new types of visualizations showing the connectivity of modes, which 
are normally depicted as isolated.  We also could continue to build on the new 
training procedure we proposed here, to find curves with particularly desirable properties,
such as diversity of networks.  Indeed, we could start to use entirely new loss 
functions, such as line and surface integrals of cross-entropy across structured
regions of weight space.

\paragraph{Acknowledgements.} Timur Garipov was supported by Ministry of Education and Science of the Russian Federation (grant 14.756.31.0001). Timur Garipov and Dmitrii Podoprikhin were supported by Samsung Research, Samsung Electronics. Andrew Gordon Wilson and Pavel Izmailov were supported by Facebook Research and NSF IIS-1563887.

\bibliographystyle{plainnat}
\bibliography{bibliography}

\begin{thebibliography}{27}
\providecommand{\natexlab}[1]{#1}
\providecommand{\url}[1]{\texttt{#1}}
\expandafter\ifx\csname urlstyle\endcsname\relax
  \providecommand{\doi}[1]{doi: #1}\else
  \providecommand{\doi}{doi: \begingroup \urlstyle{rm}\Url}\fi

\bibitem[Abadi et~al.(2016)Abadi, Agarwal, Barham, Brevdo, Chen, Citro,
  Corrado, Davis, Dean, Devin, et~al.]{abadi2016tensorflow}
Mart{\'\i}n Abadi, Ashish Agarwal, Paul Barham, Eugene Brevdo, Zhifeng Chen,
  Craig Citro, Greg~S Corrado, Andy Davis, Jeffrey Dean, Matthieu Devin, et~al.
\newblock Tensorflow: Large-scale machine learning on heterogeneous distributed
  systems.
\newblock \emph{arXiv preprint arXiv:1603.04467}, 2016.

\bibitem[Auer et~al.(1996)Auer, Herbster, and Warmuth]{auer1996exponentially}
Peter Auer, Mark Herbster, and Manfred~K Warmuth.
\newblock Exponentially many local minima for single neurons.
\newblock In \emph{Advances in Neural Information Processing Systems}, pages
  316--322, 1996.

\bibitem[Choromanska et~al.(2015)Choromanska, Henaff, Mathieu, Arous, and
  LeCun]{choromanska2015}
Anna Choromanska, Mikael Henaff, Michael Mathieu, G{\'e}rard~Ben Arous, and
  Yann LeCun.
\newblock The loss surfaces of multilayer networks.
\newblock In \emph{Artificial Intelligence and Statistics}, pages 192--204,
  2015.

\bibitem[Dauphin et~al.(2014)Dauphin, Pascanu, Gulcehre, Cho, Ganguli, and
  Bengio]{dauphin2014identifying}
Yann~N Dauphin, Razvan Pascanu, Caglar Gulcehre, Kyunghyun Cho, Surya Ganguli,
  and Yoshua Bengio.
\newblock Identifying and attacking the saddle point problem in
  high-dimensional non-convex optimization.
\newblock In \emph{Advances in Neural Information Processing Systems}, pages
  2933--2941, 2014.

\bibitem[Dinh et~al.(2017)Dinh, Pascanu, Bengio, and Bengio]{pmlr-v70-dinh17b}
Laurent Dinh, Razvan Pascanu, Samy Bengio, and Yoshua Bengio.
\newblock Sharp minima can generalize for deep nets.
\newblock In Doina Precup and Yee~Whye Teh, editors, \emph{Proceedings of the
  34th International Conference on Machine Learning}, volume~70 of
  \emph{Proceedings of Machine Learning Research}, pages 1019--1028,
  International Convention Centre, Sydney, Australia, 06--11 Aug 2017. PMLR.
\newblock URL \url{http://proceedings.mlr.press/v70/dinh17b.html}.

\bibitem[Draxler et~al.(2018)Draxler, Veschgini, Salmhofer, and
  Hamprecht]{draxler2018}
Felix Draxler, Kambis Veschgini, Manfred Salmhofer, and Fred Hamprecht.
\newblock Essentially no barriers in neural network energy landscape.
\newblock In Jennifer Dy and Andreas Krause, editors, \emph{Proceedings of the
  35th International Conference on Machine Learning}, volume~80 of
  \emph{Proceedings of Machine Learning Research}, pages 1309--1318,
  Stockholmsmässan, Stockholm Sweden, 10--15 Jul 2018. PMLR.
\newblock URL \url{http://proceedings.mlr.press/v80/draxler18a.html}.

\bibitem[Freeman and Bruna(2017)]{freeman2017topology}
C~Daniel Freeman and Joan Bruna.
\newblock Topology and geometry of half-rectified network optimization.
\newblock \emph{International Conference on Learning Representations}, 2017.

\bibitem[Goodfellow et~al.(2015)Goodfellow, Vinyals, and Saxe]{Goodfellow2015}
Ian~J Goodfellow, Oriol Vinyals, and Andrew~M Saxe.
\newblock Qualitatively characterizing neural network optimization problems.
\newblock \emph{International Conference on Learning Representations}, 2015.

\bibitem[Gotmare et~al.(2018)Gotmare, Shirish~Keskar, Xiong, and
  Socher]{gotmare2018using}
Akhilesh Gotmare, Nitish Shirish~Keskar, Caiming Xiong, and Richard Socher.
\newblock Using mode connectivity for loss landscape analysis.
\newblock \emph{arXiv preprint arXiv:1806.06977}, 2018.

\bibitem[Guo et~al.(2017)Guo, Pleiss, Sun, and Weinberger]{guo2017}
Chuan Guo, Geoff Pleiss, Yu~Sun, and Kilian~Q Weinberger.
\newblock On calibration of modern neural networks.
\newblock In \emph{International Conference on Machine Learning}, pages
  1321--1330, 2017.

\bibitem[He et~al.(2016)He, Zhang, Ren, and Sun]{he2016deep}
Kaiming He, Xiangyu Zhang, Shaoqing Ren, and Jian Sun.
\newblock Deep residual learning for image recognition.
\newblock In \emph{Proceedings of the IEEE conference on computer vision and
  pattern recognition}, pages 770--778, 2016.

\bibitem[Hochreiter and Schmidhuber(1997)]{hochreiter1997flat}
Sepp Hochreiter and J{\"u}rgen Schmidhuber.
\newblock Flat minima.
\newblock \emph{Neural Computation}, 9\penalty0 (1):\penalty0 1--42, 1997.

\bibitem[Huang et~al.(2017)Huang, Li, Pleiss, Liu, Hopcroft, and
  Weinberger]{huang2017}
Gao Huang, Yixuan Li, Geoff Pleiss, Zhuang Liu, John~E Hopcroft, and Kilian~Q
  Weinberger.
\newblock Snapshot ensembles: Train 1, get m for free.
\newblock \emph{International Conference on Learning Representations}, 2017.

\bibitem[Ioffe and Szegedy(2015)]{ioffe2015}
Sergey Ioffe and Christian Szegedy.
\newblock Batch normalization: Accelerating deep network training by reducing
  internal covariate shift.
\newblock In \emph{International Conference on Machine Learning}, pages
  448--456, 2015.

\bibitem[Izmailov et~al.(2018)Izmailov, Podoprikhin, Garipov, Vetrov, and
  Wilson]{izmailov2018averaging}
Pavel Izmailov, Dmitrii Podoprikhin, Timur Garipov, Dmitry Vetrov, and
  Andrew~Gordon Wilson.
\newblock Averaging weights leads to wider optima and better generalization.
\newblock \emph{arXiv preprint arXiv:1803.05407}, 2018.

\bibitem[Jonsson et~al.(1998)Jonsson, Mills, and Jacobsen]{jonson1998}
Hannes Jonsson, Greg Mills, and Karsten~W Jacobsen.
\newblock {Nudged elastic band method for finding minimum energy paths of
  transitions.}
\newblock \emph{Classical and quantum dynamics in condensed phase simulations},
  1998.

\bibitem[Keskar et~al.(2017)Keskar, Mudigere, Nocedal, Smelyanskiy, and
  Tang]{keskar2017large}
Nitish~Shirish Keskar, Dheevatsa Mudigere, Jorge Nocedal, Mikhail Smelyanskiy,
  and Ping Tak~Peter Tang.
\newblock On large-batch training for deep learning: Generalization gap and
  sharp minima.
\newblock \emph{International Conference on Learning Representations}, 2017.

\bibitem[Lee et~al.(2016{\natexlab{a}})Lee, Simchowitz, Jordan, and
  Recht]{lee2016gradient}
Jason~D Lee, Max Simchowitz, Michael~I Jordan, and Benjamin Recht.
\newblock Gradient descent only converges to minimizers.
\newblock In \emph{Conference on Learning Theory}, pages 1246--1257,
  2016{\natexlab{a}}.

\bibitem[Lee et~al.(2016{\natexlab{b}})Lee, Prakash, Cogswell, Ranjan,
  Crandall, and Batra]{lee2016stochastic}
Stefan Lee, Senthil Purushwalkam~Shiva Prakash, Michael Cogswell, Viresh
  Ranjan, David Crandall, and Dhruv Batra.
\newblock Stochastic multiple choice learning for training diverse deep
  ensembles.
\newblock In \emph{Advances in Neural Information Processing Systems}, pages
  2119--2127, 2016{\natexlab{b}}.

\bibitem[Li et~al.(2017)Li, Xu, Taylor, and Goldstein]{li2017visualizing}
Hao Li, Zheng Xu, Gavin Taylor, and Tom Goldstein.
\newblock Visualizing the loss landscape of neural nets.
\newblock \emph{arXiv preprint arXiv:1712.09913}, 2017.

\bibitem[Loshchilov and Hutter(2017)]{loshchilov2017}
Ilya Loshchilov and Frank Hutter.
\newblock Sgdr: Stochastic gradient descent with warm restarts.
\newblock \emph{International Conference on Learning Representations}, 2017.

\bibitem[Marcus et~al.(1993)Marcus, Marcinkiewicz, and
  Santorini]{marcus1993building}
Mitchell~P Marcus, Mary~Ann Marcinkiewicz, and Beatrice Santorini.
\newblock Building a large annotated corpus of english: The penn treebank.
\newblock \emph{Computational linguistics}, 19\penalty0 (2):\penalty0 313--330,
  1993.

\bibitem[Russakovsky et~al.(2012)Russakovsky, Deng, Su, Krause, Satheesh, Ma,
  Huang, Karpathy, Khosla, Bernstein, et~al.]{russakovsky2015imagenet}
Olga Russakovsky, Jia Deng, Hao Su, Jonathan Krause, Sanjeev Satheesh, Sean Ma,
  Zhiheng Huang, Andrej Karpathy, Aditya Khosla, Michael Bernstein, et~al.
\newblock Imagenet large scale visual recognition challenge.
\newblock \emph{International Journal of Computer Vision}, 115\penalty0
  (3):\penalty0 211--252, 2012.

\bibitem[Simonyan and Zisserman(2014)]{simonyan2014very}
Karen Simonyan and Andrew Zisserman.
\newblock Very deep convolutional networks for large-scale image recognition.
\newblock \emph{arXiv preprint arXiv:1409.1556}, 2014.

\bibitem[Smith and Topin(2017)]{smith2017exploring}
Leslie~N Smith and Nicholay Topin.
\newblock Exploring loss function topology with cyclical learning rates.
\newblock \emph{arXiv preprint arXiv:1702.04283}, 2017.

\bibitem[Xie et~al.(2013)Xie, Xu, and Chuang]{xie2013horizontal}
Jingjing Xie, Bing Xu, and Zhang Chuang.
\newblock Horizontal and vertical ensemble with deep representation for
  classification.
\newblock \emph{arXiv preprint arXiv:1306.2759}, 2013.

\bibitem[Zagoruyko and Komodakis(2016)]{zagoruyko2016wide}
Sergey Zagoruyko and Nikos Komodakis.
\newblock Wide residual networks.
\newblock In \emph{BMVC}, 2016.

\end{thebibliography}

\appendix
\section{Supplementary Material}
We organize the supplementary material  as follows. Section \ref{sec:computational_complexity} 
discusses the computational complexity of the 
proposed curve finding method. Section \ref{sec:batchnorm} describes how to apply batch normalization at test time to points on curves connecting pairs of local optima.
Section \ref{sec:formulas_n_bends} provides formulas for a
polygonal chain and Bezier curve with $n$ bends.
Section \ref{sec:curve_exp_res} provides details and results of 
experiments on curve finding and contains a table summarizing all path finding experiments.
Section \ref{sec:surfaces} provides additional visualizations of the train loss
and test accuracy surfaces.
Section \ref{sec:curve_ensembling_exp} contains details on curve ensembling experiments. Section \ref{sec:curve_understanding_exp} describes experiments on relation between mode connectivity and the number of parameters in the networks.
Section \ref{sec:trivial_curves}
discusses a trivial construction of curves connecting two modes, where 
points on the curve represent reparameterization of the endpoints, unlike
the curves in the main text. Section \ref{sec:fge_exp_details} provides details of experiments on FGE. Finally, Section \ref{sec:fge_curves} describes pathways traversed by FGE.

\subsection{Computational complexity of curve finding}
\label{sec:computational_complexity}

The forward pass of the proposed method consists of two steps: computing the point
$\phi_\theta(t)$ and then passing a mini-batch of data through the DNN corresponding
to this point. Similarly, the backward pass consists of first computing the 
gradient of the loss with respect to $\phi_\theta(t)$, and then multiplying the result
by the Jacobian $\frac {\partial \phi_\theta}{\partial \theta}$. The second step of
the forward pass and the first step of the backward pass are exactly the same as
the forward and backward pass in the training of a single DNN model. The additional
computational complexity of the procedure compared to single model training comes
from the first step of the forward pass and the second step of the backward pass and
in general depends on the parametrization $\phi_\theta(\cdot)$ of the curve.

In our experiments we use curve parametrizations of a specific form. The general
formula for a curve with one bend is given by
\[
\phi_\theta(t) = \hat w_1 \cdot c_1(t) + \theta \cdot c(t) + \hat w_2 \cdot c_2(t).
\]
Here the parameters of the curve are given by $\theta \in \mathbb{R}^{|net|}$ and coefficients $c_1, c_2, c: [0,1] \to \mathbb{R}$.

For this family of curves the computational complexity of the first step of the method
is $\mathcal{O}(|net|)$, as we only need to compute a weighted sum of $\hat w_1$, 
$\hat w_2$ and
$\theta \in \mathbb{R}^{|net|}$. The Jacobian matrix
\[
\frac{\partial \phi_\theta(t)}{\partial \theta} = c(t) \cdot I,
\]
thus the additional computational complexity of the backward pass is also 
$\mathcal{O}(|net|)$, as we only need to multiply the gradient with respect to 
$\phi_\theta (t)$ by a scalar. Thus, the total additional computational complexity 
is $\mathcal{O}(|net|)$. In practice we observe that the gap in time-complexity 
between one epoch of training a single model and one epoch of the proposed method
with the same network architecture is usually below $50 \%$.

\subsection{Batch Normalization}
\label{sec:batchnorm}
Batch normalization (\citet{ioffe2015}) is essential to modern deep learning 
architectures. Batch normalization re-parametrizes the output of each layer as
\[
\hat x = \gamma \frac{x - \mu(x)}{\sigma(x) + \epsilon} + \beta,
\]
where $\mu(x)$ and $\sigma(x)$ are the mean and standard deviation of the 
output $x$, $\epsilon > 0$ is a constant for numerical stability
and $\gamma$ and 
$\beta$ are free parameters. 
During training, $\mu(x)$ and $\sigma(x)$ are computed 
separately for each mini-batch and at test time statistics aggregated during training are used.

When connecting two DNNs that use batch normalization, along a curve 
$\phi(t)$, we compute $\mu(x)$ and $\sigma(x)$ for any given $t$ over 
mini-batches during training,
as usual.  In order to apply batch-normalization to a network on the curve 
at the test stage we compute these statistics with one additional
pass over the data, as running averages for these
networks are not collected during training.

\subsection{Formulas for curves with $n$ bends}

\label{sec:formulas_n_bends}

For $n$ bends $\theta = \{w_1, w_2, \ldots, w_n\}$, the parametrization of a
polygonal chain connecting points $w_0, w_{n+1}$ is given by
\[
\phi_{\theta}(t) = 
(n+1) \cdot \left(\left(t - \frac{i}{n + 1}\right) \cdot w_{i+1} +  \left(\frac{i+1}{n + 1} - t\right) \cdot w_i\right), 
\]
for $\frac{i}{n + 1} \le t \le \frac{i+1}{n+1}$ and $0 \leq i \leq n$.

For $n$ bends $\theta = \{w_1, w_2, \ldots, w_n\}$, the parametrization of a
Bezier curve connecting points $w_0$ and $w_{n+1}$ is given by
\[
\phi_{\theta}(t) = \sum\limits_{i = 0}^{n+1}w_i C_{n + 1}^{i} t^{i} (1 - t)^{n +1 - i}
\]

\subsection{Curve Finding Experiments}
\label{sec:curve_exp_res}

\begin{table*}[!t]
	\begin{adjustwidth}{-1in}{-1in}
	\centering
	\caption{The properties of loss and error values along the found curves for different architectures and tasks}
	\label{table:curves}
	\begin{tabular}{ll  l  llll  lll  lll}
		\toprule
		\multicolumn{2}{c}{Model}& \multicolumn{1}{c}{Length} & \multicolumn{4}{c}{Train Loss} & \multicolumn{3}{c}{Train Error (\%)}
		& \multicolumn{3}{c}{Test Error (\%)} 
		\\
		\cmidrule(lr){1-2}
	    \cmidrule(lr){3-3}
 	    \cmidrule(lr){4-7}
   	    \cmidrule(lr){8-10}
   	    \cmidrule(lr){11-13}
		DNN & Curve
		& Ratio
		& Min & Int & Mean & Max
		& Min & Int & Max
		& Min & Int & Max
		\\
		\midrule
		\multicolumn{13}{c}{MNIST}\\
		\midrule
		FC & Single
		& ---
		& $0.018$ & --- & --- & $0.018$
		& $0.01$  & --- & $0.01$
		& $1.46$  & --- & $1.5$
		\\
		FC & Segment
		& $1$
		& $0.018$  & $0.252$ & $0.252$ & $0.657$
		& $0.01$   & $0.53$   &	$2.13$
		& $1.45$   & $1.96$   &	$3.18$
		\\
		FC & Bezier
		& $1.58$
		& $0.016$ & $0.02$ & $0.02$ & $0.024$
		& $0.01$ & $0.02$ & $0.04$
		& $1.46$ & $1.52$  & $1.56$
		\\
		FC & Polychain
		& $1.73$
		& $0.013$ & $0.022$ & $0.022$ & $0.029$
		& $0$	   & $0.03$	  & $0.07$
		& $1.46$   & $1.51$   &	$1.58$
		\\
		\midrule
		\multicolumn{13}{c}{CIFAR-10}\\
		\midrule
		3conv3fc & Single
		& ---
		& $0.05$ & --- & --- & $0.05$
		& $0.06$ & --- & $0.06$
		& $12.3$ & --- & $12.36$
		\\
		3conv3fc & Segment
		& $1$
		& $0.05$ & $1.124$ & $1.124$ & $2.416$
		& $0.06$  & $35.69$	& $88.24$
		& $12.28$ & $43.3$	& $88.27$
		\\
		3conv3fc & Bezier
		& $1.30$
		& $0.034$ & $0.038$ & $0.037$ & $0.05$
		& $0.05$  & $0.1$	& $0.2$
		& $12.06$ & $12.7$	& $13.66$
		\\
		3conv3fc & Polychain
		& $1.67$	
		& $0.04$  & $0.044$ & $0.044$ & $0.05$
		& $0.06$  & $0.15$  & $0.31$
		& $12.17$ & $12.68$ & $13.31$
		\\
		\midrule
		VGG-16 & Single
		& ---
		& $0.04$ & --- & --- & $0.04$
		& $0$	 & --- & $0.01$
		& $6.87$ & --- & $7.01$
		\\
		VGG-16 & Segment
		& $1$
		& $0.039$ & $1.759$ & $1.759$ & $2.569$
		& $0$	  & $61.43$ & $90$
		& $6.87$  & $63.75$ & $90$
		\\
		VGG-16 & Bezier
		& $1.55$
		& $0.028$ & $0.03$ & $0.03$ & $0.04$
		& $0$     & $0.01$ & $0.02$
		& $6.59$  & $6.77$ & $7.01$
		\\
		VGG-16 & Polychain
		& $1.83$
		& $0.025$ & $0.031$ & $0.031$ & $0.045$
		& $0$     & $0.01$  & $0.04$
		& $6.54$  & $6.89$	& $7.28$
		\\
		\midrule
		ResNet-158 & Single
		& ---
		& $0.015$ & --- & --- & $0.015$
		& $0.02$  & --- & $0.02$
		& $5.56$  & --- & $5.74$
		\\
		ResNet-158 & Segment
		& $1$
		& $0.013$ & $0.551$ & $0.551$ & $2.613$
		& $0$     & $16.37$ & $81.41$
		& $5.57$  & $20.79$ & $80.00$
		\\
		ResNet-158 & Bezier 
		& $2.13$
		& $0.013$ & $0.017$ & $0.018$ & $0.022$
		& $0$  & $0.02$	& $0.07$
		& $5.48$  & $5.82$	& $6.24$
		\\
		ResNet-158 & Polychain
		& $3.48$
		& $0.013$ & $0.017$ & $0.017$ & $0.047$
		& $0$     & $0.05$	& $0.139$
		& $5.48$  & $5.88$	& $7.35$
		\\
		\midrule
		WRN-10-28 & Single
		& ---
		& $0.033$ & --- & --- & $0.035$
		& $0$     & --- & $0$
		& $4.49$  & --- & $4.56$
		\\
		WRN-10-28 & Segment
		& $1$
		& $0.033$ & $0.412$ & $0.412$ & $2.203$
		& $0$	  & $5.44$  & $65.62$
		& $4.49$  & $10.55$ & $66.6$
		\\
		WRN-10-28 & Bezier
		& $1.83$
		& $0.03$  & $0.033$ & $0.038$ & $0.038$
		& $0$     & $0.01$  & $0.04$
		& $4.4$   & $4.62$	& $4.83$
		\\
		WRN-10-28 & Polychain
		& $1.95$
		& $0.026$ & $0.029$ & $0.029$ & $0.037$
		& $0$     & $0$ 	& $0$
		& $4.38$  & $6.93$  & $10.38$
		\\
		\midrule
		\multicolumn{13}{c}{CIFAR-100}\\
		\midrule
		VGG-16 & Single
		& ---
		& $0.14$  & --- & --- & $0.141$
		& $0.05$  & --- & $0.06$
		& $29.44$ & --- & $29.94$
		\\
		VGG-16 & Segment
		& $1$    
		& $0.137$ & $3.606$ & $3.606$ & $4.941$
		& $0.04$  & $73.25$ & $99$
		& $29.44$ & $80.59$ & $99.01$
		\\
		VGG-16 & Bezier
		& $1.52$
		& $0.095$ & $0.107$ & $0.105$ & $0.141$
		& $0.03$  & $0.08$  & $0.18$
		& $29.28$ & $30.49$ & $31.23$
		\\
		VGG-16 & Polychain
		& $1.64$
		& $0.118$ & $0.139$ & $0.139$ & $0.2$
		& $0.04$  & $0.19$	& $0.39$
		& $29.33$ & $30.13$ & $30.92$
		\\
		\midrule
		ResNet-164 & Single
		& ---
		& $0.079$ & --- & --- & $0.08$
		& $0.06$  & --- & $0.09$
		& $24.41$ & --- & $24.4$
		\\
		ResNet-164 & Segment
		& $1$    
		& $0.076$ & $1.844$ & $1.844$ & $5.53$
		& $0.06$  & $38.03$ & $98.65$
		& $24.4$  & $53.69$ & $98.83$
		\\
		ResNet-164 & Bezier
		& $1.87$
		& $0.074$ & $0.083$ & $0.084$ & $0.098$
		& $0.05$  & $0.28$  & $0.96$
		& $24.15$ & $24.99$ & $26.1$
		\\
		ResNet-164 & Polychain
		& $2.56$
		& $0.067$ & $0.078$ & $0.078$ & $0.109$
		& $0.06$  & $0.28$  & $0.85$
		& $23.98$ & $24.92$ & $26.12$
		\\
		\bottomrule
	\end{tabular}
	\end{adjustwidth}
\end{table*}

\begin{table*}[!t]
	\caption{The value of perplexity  along the found curves for PTB dataset}
	\label{table:curves_lstm}
	\centering
	\begin{tabular}{ll  ll  ll  ll}
		\toprule
		\multicolumn{2}{c}{Model} & \multicolumn{2}{c}{Train } & \multicolumn{2}{c}{Validation}
		& \multicolumn{2}{c}{Test} 
		\\
		\cmidrule(lr){1-2}
		\cmidrule(lr){3-4}
		\cmidrule(lr){5-6}
		\cmidrule(lr){7-8}
		DNN & Curve
		& Min & Max
		& Min & Max
		& Min & Max
		\\
		\midrule
		\multicolumn{8}{c}{\small PTB}\\
		\midrule
		RNN & Single
		& $37.5$ & $39.2$ 
		& $82.7$ & $83.1$
		& $78.7$ & $78.9$
		\\
		RNN & Segment 
		& $37.5$ & $596.3$ 
		& $82.7$ & $682.1$
		& $78.7$ & $615.7$
		\\
		RNN & Bezier 
		& $29.8$ & $39.2$ 
		& $82.7$ & $88.7$
		& $78.7$ & $84.0$
		\\
		\bottomrule
	\end{tabular}
\end{table*}

\begin{figure*}[!h]
	\centering
	\begin{subfigure}{0.32\textwidth}
		\includegraphics[width=\textwidth]{pics/c100_resnet_i_3.pdf}
	\end{subfigure}
	~
	\begin{subfigure}{0.32\textwidth}
		\includegraphics[width=\textwidth]{pics/c100_resnet_b_3.pdf}
	\end{subfigure}
	~
	\begin{subfigure}{0.32\textwidth}
		\includegraphics[width=\textwidth]{pics/c100_resnet_p_3.pdf}
	\end{subfigure}
	\begin{subfigure}{0.32\textwidth}
		\includegraphics[width=\textwidth]{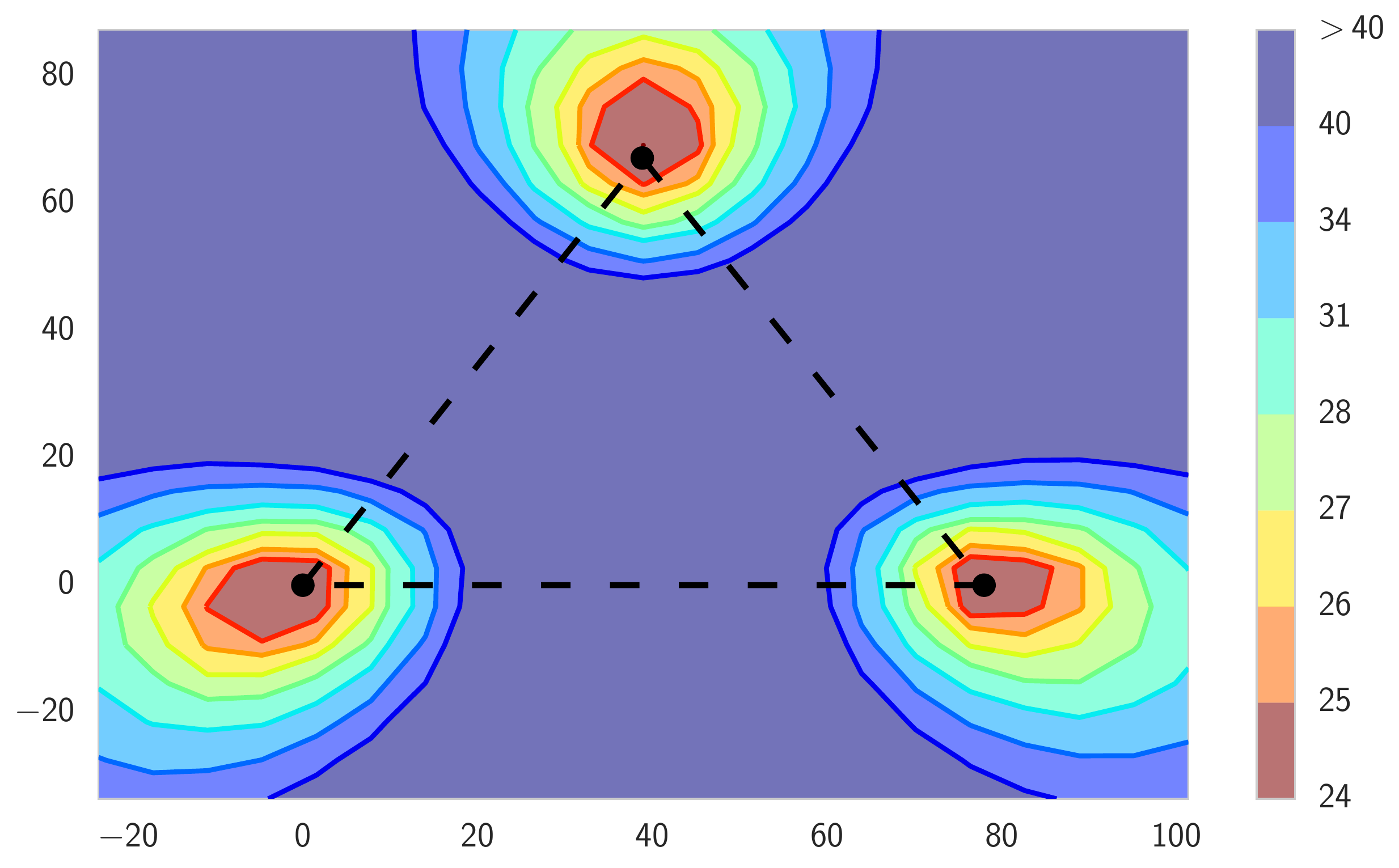}
	\end{subfigure}
	~
	\begin{subfigure}{0.32\textwidth}
		\includegraphics[width=\textwidth]{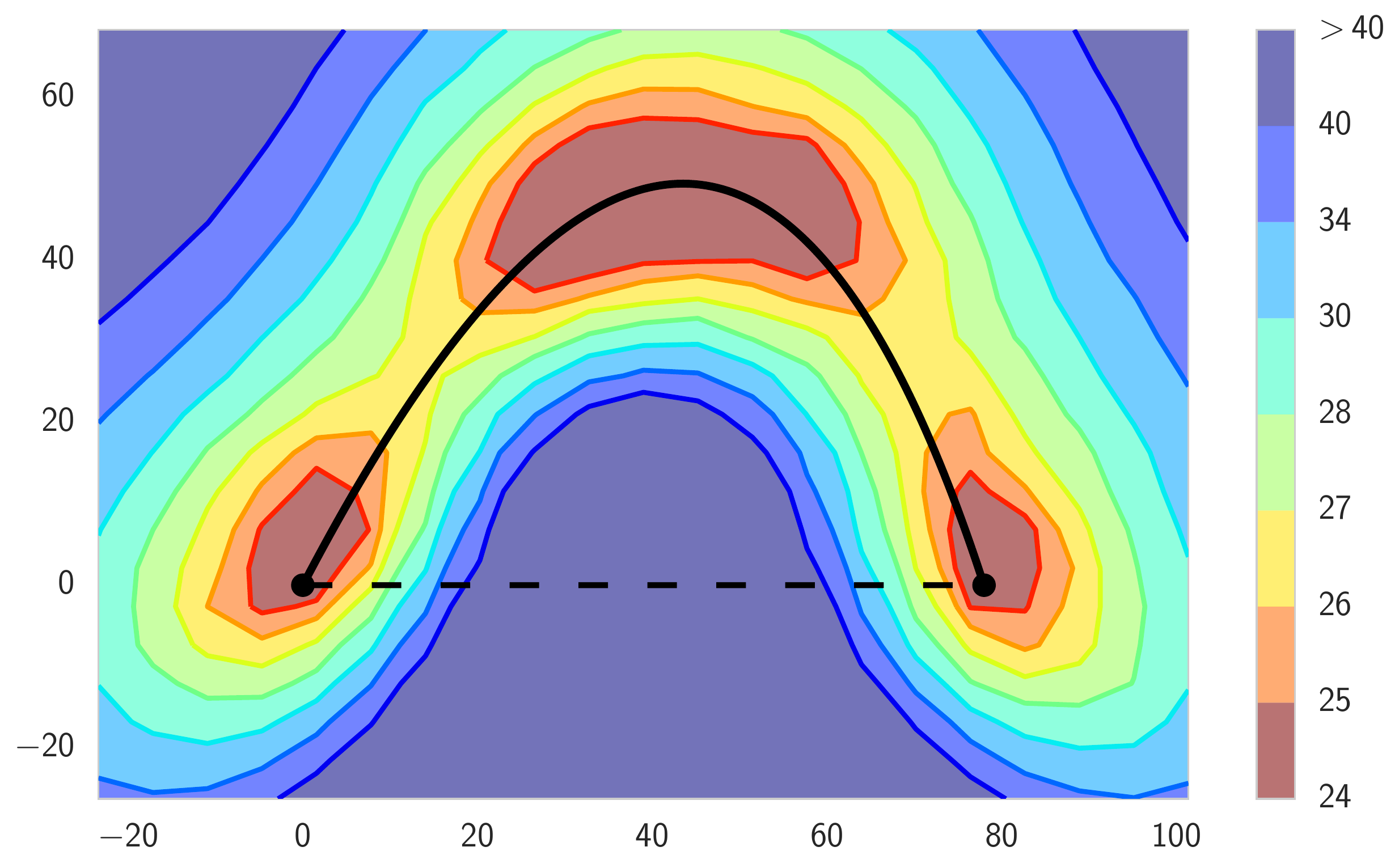}
	\end{subfigure}
	~
	\begin{subfigure}{0.32\textwidth}
		\includegraphics[width=\textwidth]{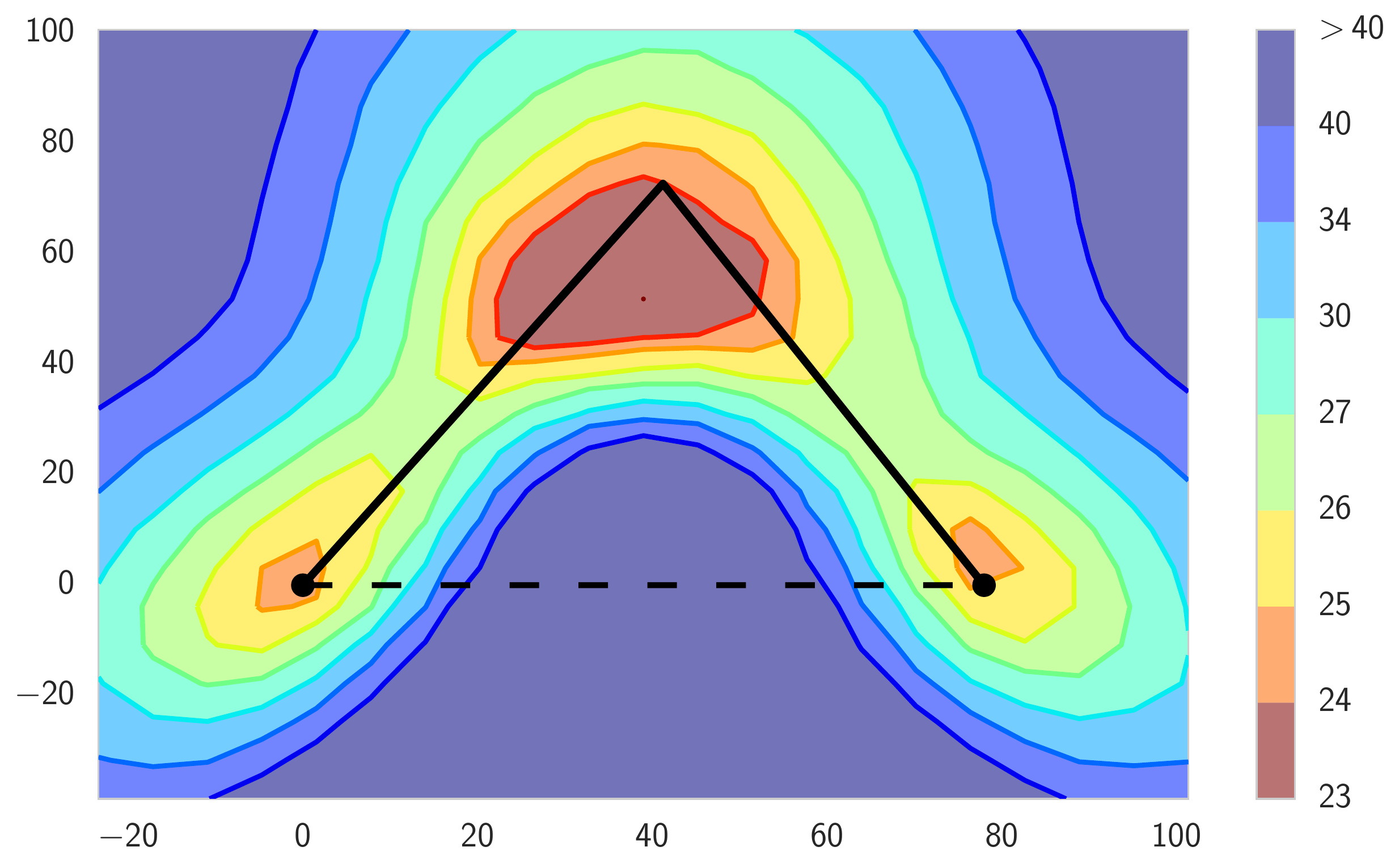}
	\end{subfigure}
	\caption{
		The $\ell_2$-regularized cross-entropy train loss (\textbf{Top}) and test error
		(\textbf{Bottom})
		surfaces of a deep residual network (ResNet-164)
		on CIFAR-100. \textbf{Left:} Three optima for independently trained networks.
		\textbf{Middle} and \textbf{Right}: A quadratic Bezier curve, and a polygonal
		chain with one bend, connecting the lower two optima on the left panel along
		a path of near-constant loss.  Notice that in each panel, a direct 
		linear path between each mode would incur high loss.}
	\label{fig:surface_resnet}
\end{figure*}

\begin{figure*}[!h]
	\centering
	\begin{subfigure}{0.32\textwidth}
		\includegraphics[width=\textwidth]{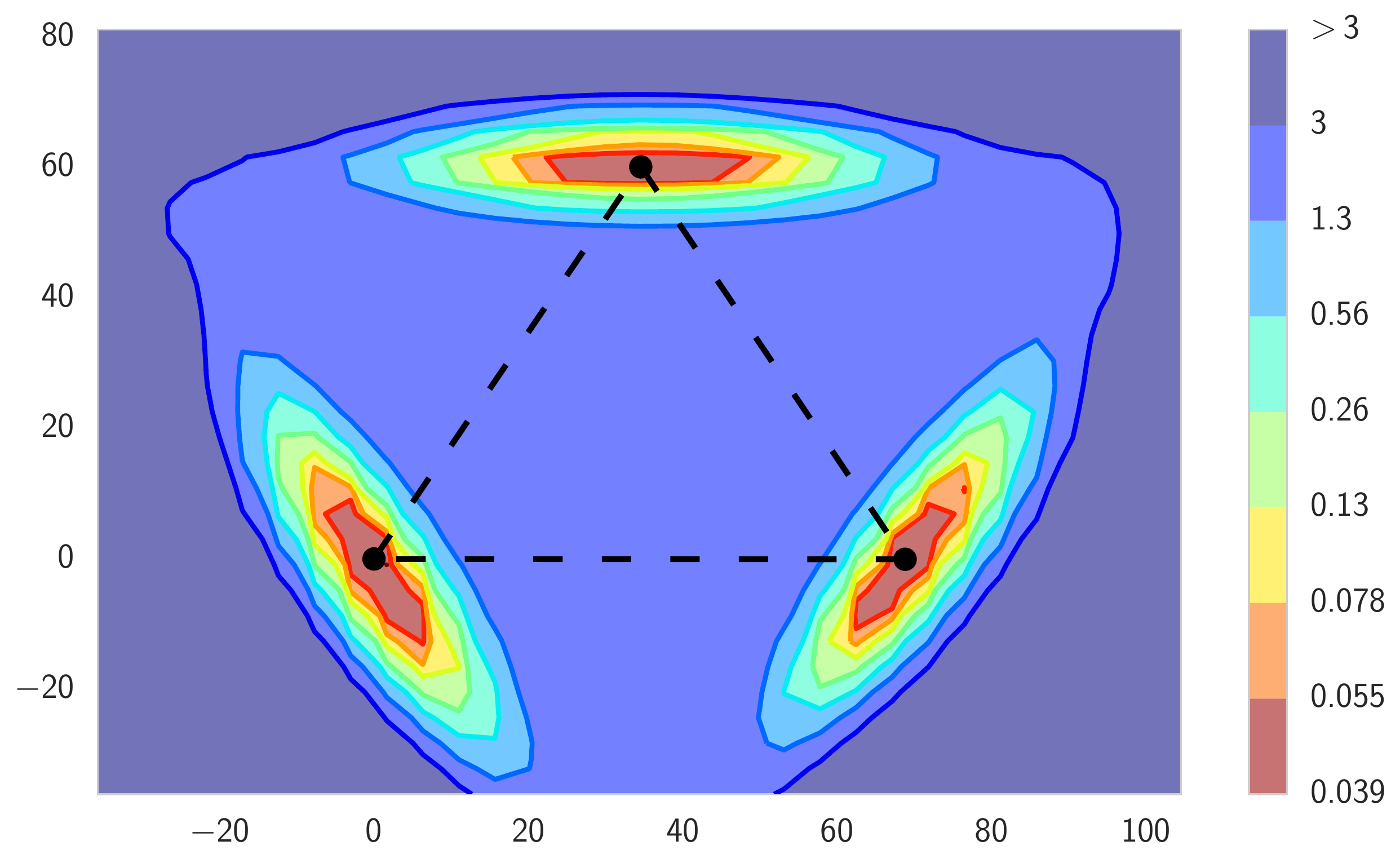}
	\end{subfigure}
	~
	\begin{subfigure}{0.32\textwidth}
		\includegraphics[width=\textwidth]{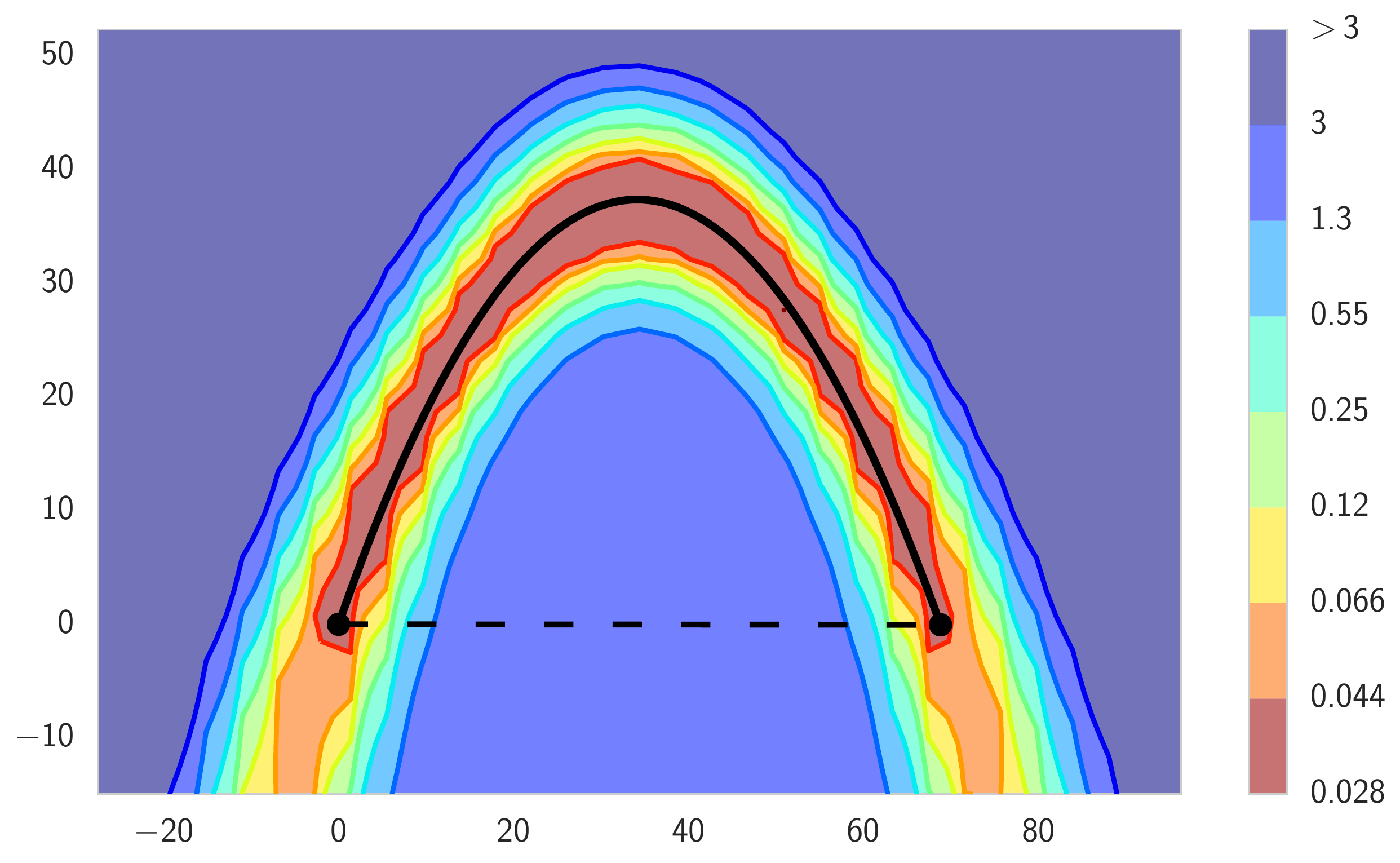}
	\end{subfigure}
	~
	\begin{subfigure}{0.32\textwidth}
		\includegraphics[width=\textwidth]{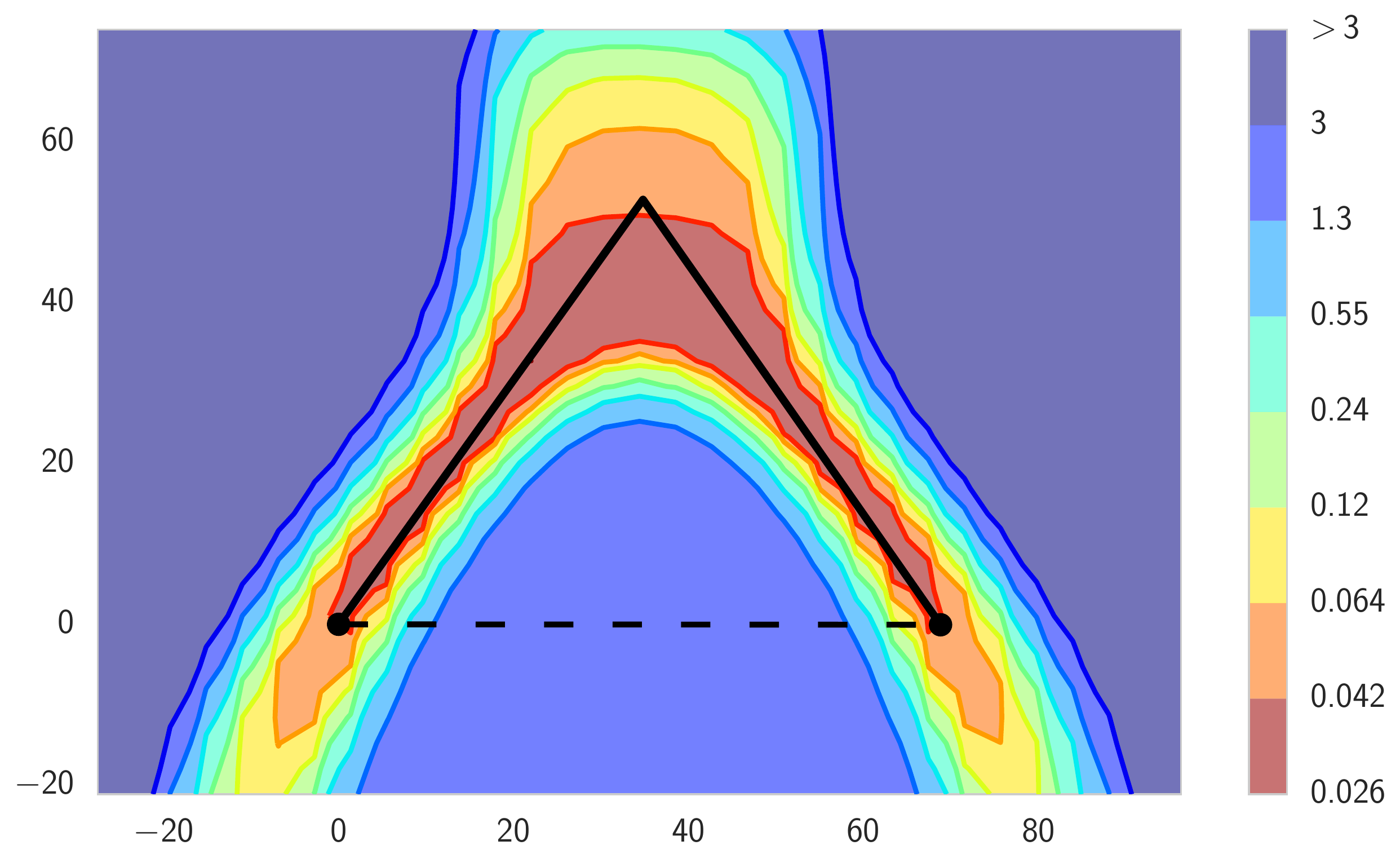}
	\end{subfigure}
	
	\begin{subfigure}{0.32\textwidth}
		\includegraphics[width=\textwidth]{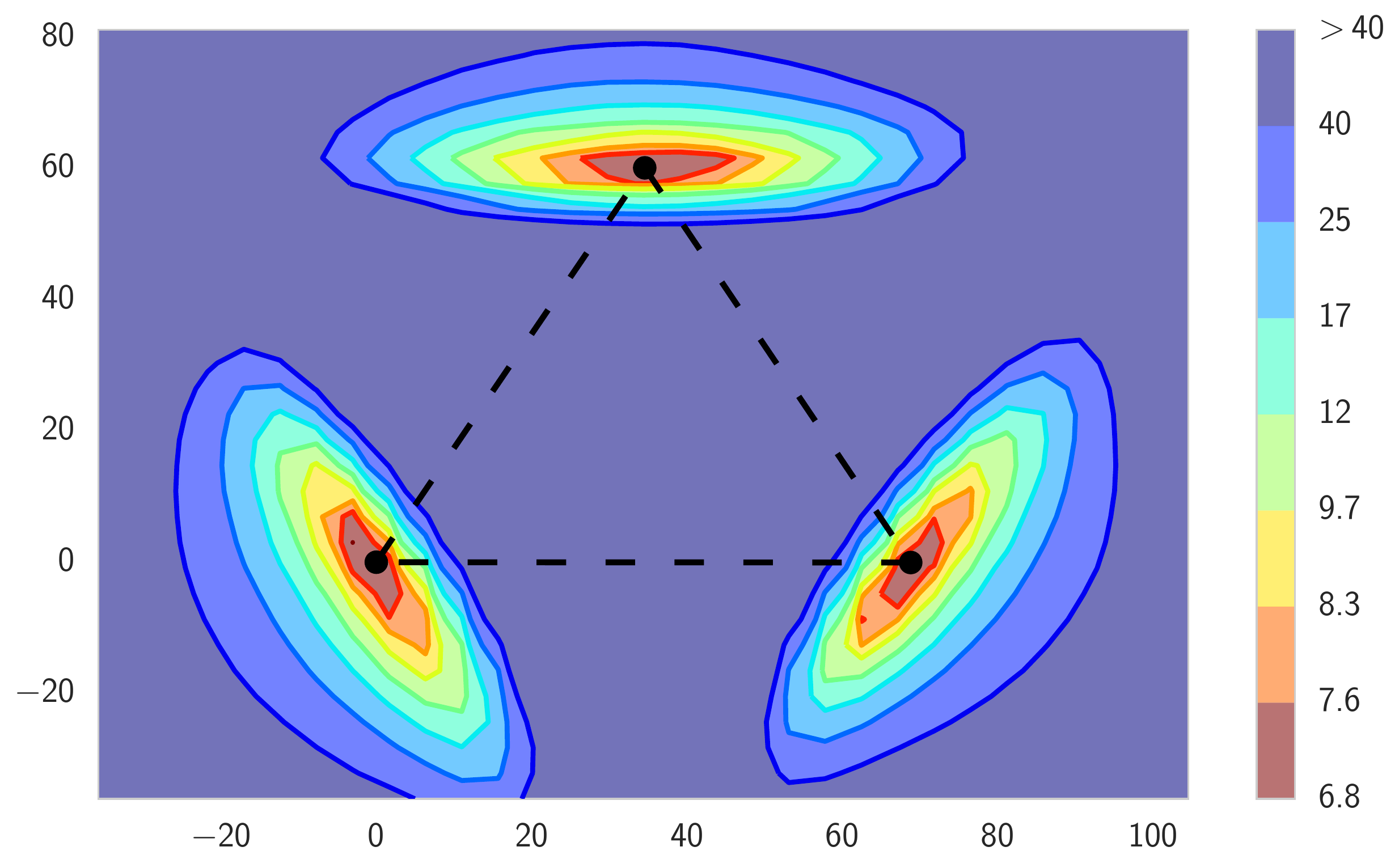}
	\end{subfigure}
	~
	\begin{subfigure}{0.32\textwidth}
		\includegraphics[width=\textwidth]{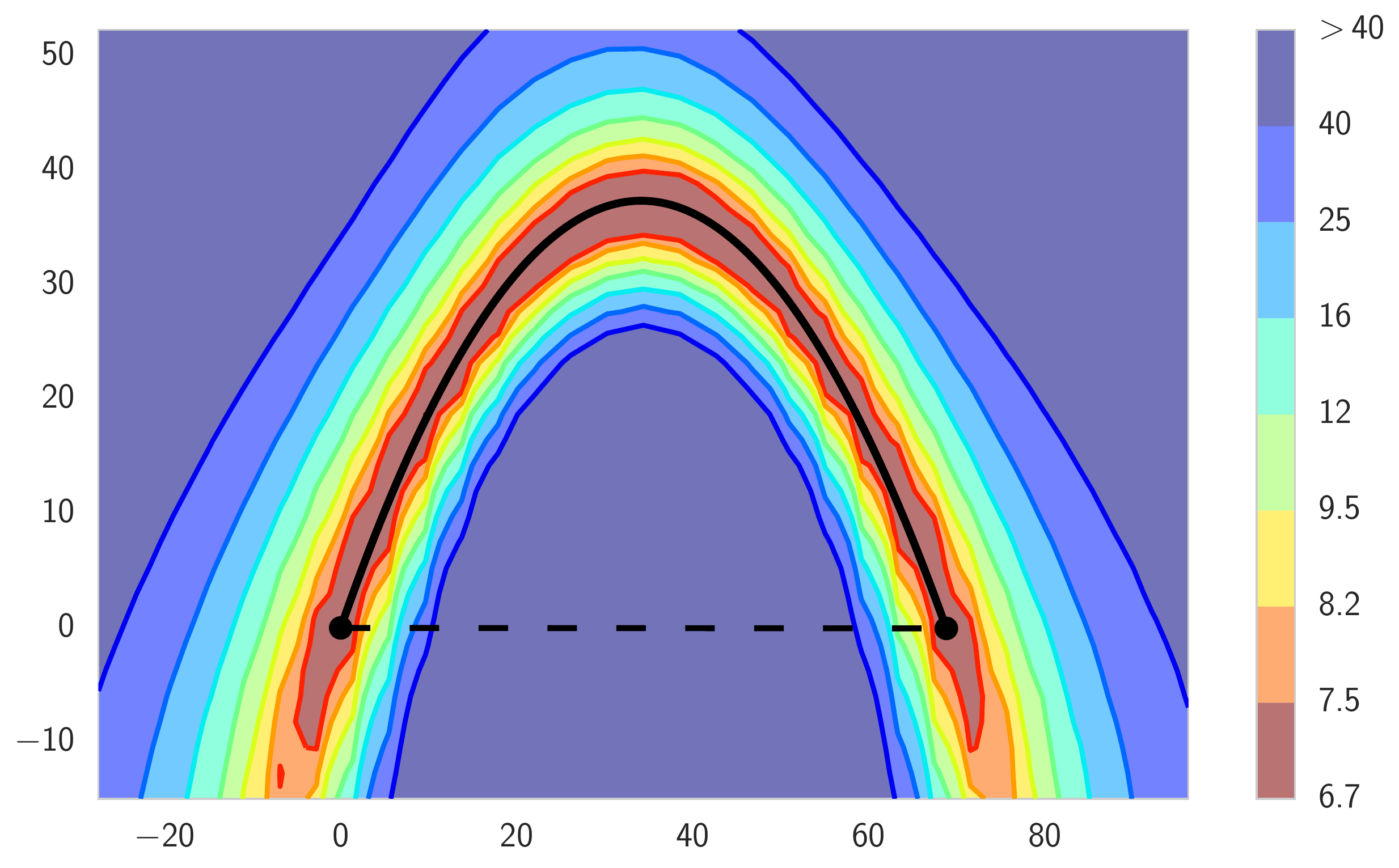}
	\end{subfigure}
	~
	\begin{subfigure}{0.32\textwidth}
		\includegraphics[width=\textwidth]{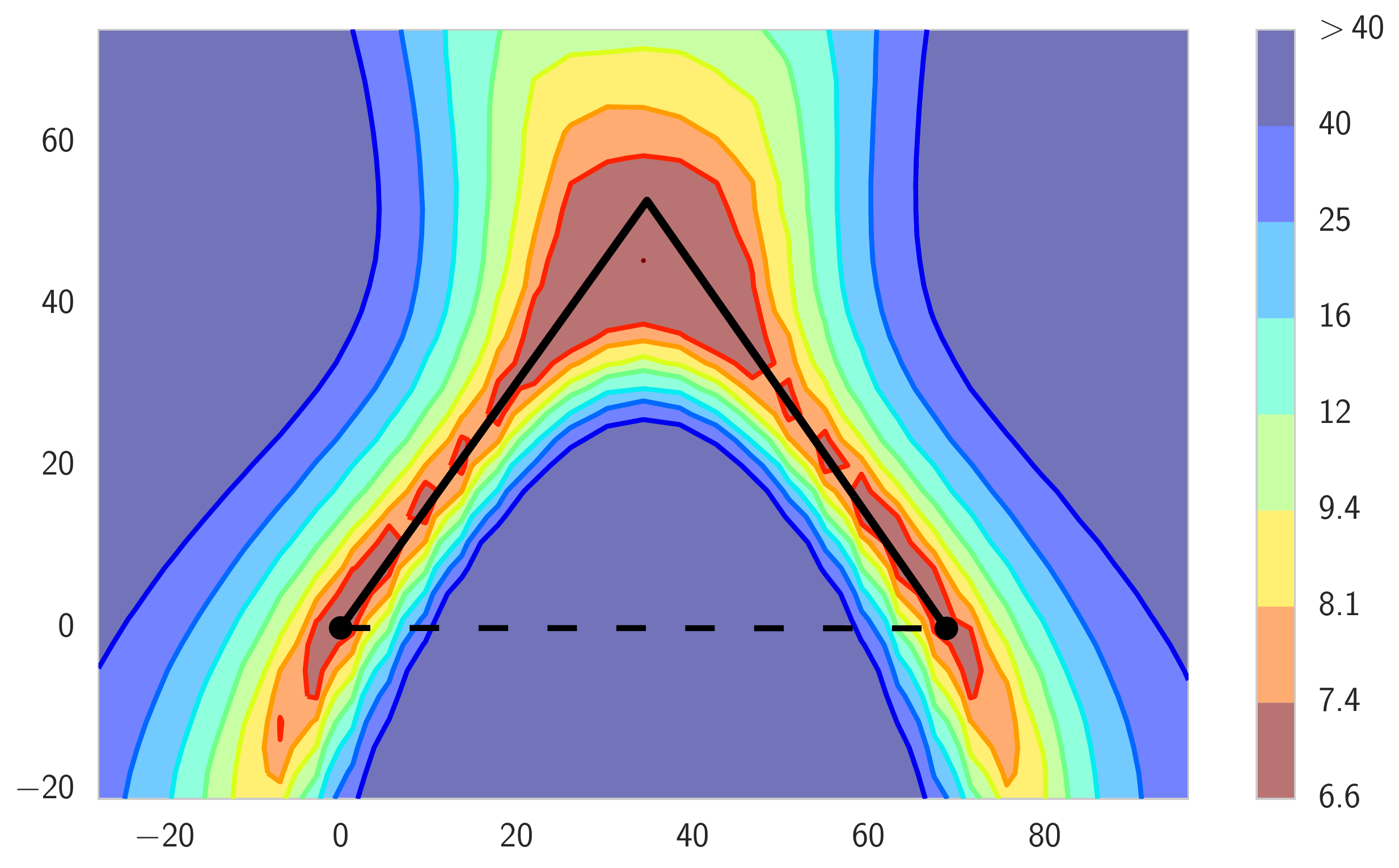}
	\end{subfigure}
	\caption{
		Same as Fig. \ref{fig:surface_resnet} for VGG-16 on CIFAR-10.
	}
	\label{fig:surface_vgg}
\end{figure*}

All experiments on curve finding were conducted with TensorFlow (\citet{abadi2016tensorflow}) 
and as baseline models we used the following implementations:
\begin{itemize}
	\item ResNet-bottleneck-$164$ and Wide ResNet-28-10 (\url{https://github.com/tensorflow/models/tree/master/research/resnet});
	\item ResNet-158 (\url{https://github.com/tensorflow/models/tree/master/official/resnet});
	
	\item A reimplementation of VGG-16  without batch-normalization from (\url{https://github.com/pytorch/vision/blob/master/torchvision/models/vgg.py});
\end{itemize}

Table \ref{table:curves} summarizes the results of the 
curve finding experiments with 
all datasets and architectures.
For each of the models we report the properties of loss and the 
error on the train and test datasets. For 
each of these metrics we report 3 values: ``Max'' is the maximum values 
of the metric along the curve, ``Int'' is a numerical approximation of the integral
$
{\int {\mbox{$<$metric$>$}(\phi_\theta)} d\phi_\theta} / {\int d\phi_\theta},
$
where $<$metric$>$ represents the train loss or the error on the train or test
dataset and ``Min'' is the minimum value of the error on the
curve. ``Int''
represents a mean over a uniform distribution on the curve, and for the train
loss it  \textbf{coincides with the loss ($1$)} in the paper. 
We use an equally-spaced grid with $121$ points 
on $[0, 1]$ to estimate the values of ``Min'', ``Max'', ``Int''. 
For ``Int'' we use the trapezoidal rule to estimate the integral. 
For each dataset and architecture we report the performance of single models
used as the endpoints of the curve as ``Single'', the performance of a line
segment connecting the two single networks as ``Segment'', the
performance of a quadratic Bezier curve as ``Bezier'' and the performance of a polygonal chain with one bend as
``Polychain''. Finally, for each curve we report the ratio of its length to the
length of a line segment connecting the two modes.

We also examined the quantity ``Mean'' defined as 
$\int {<\mbox{metric}>(\phi_\theta (t))} dt$,  which \textbf{coincides with the loss
($2$)} from the paper, but in all our experiments it is nearly  
equal to ``Int''.

Besides convolutional and fully-connected architectures we also apply our 
approach to RNN architecture on next word prediction task, PTB dataset 
(\citet{marcus1993building}). As a base model we used the implementation  
available at \url{https://www.tensorflow.org/tutorials/recurrent}.
As the main loss we consider  perplexity. The results are presented in Table
\ref{table:curves_lstm}.

\subsection{Train loss and test accuracy surfaces}
\label{sec:surfaces}

In this section we provide additional visualizations.
Fig. \ref{fig:surface_resnet} and Fig. \ref{fig:surface_vgg} 
show visualizations of the train loss and test accuracy for
ResNet-$164$ on CIFAR-100 and VGG-$16$ on CIFAR-10.

\subsection{Curve Ensembling}
\label{sec:curve_ensembling_exp}
\begin{figure*}[!h]
	\centering
	\begin{subfigure}{0.46\textwidth}
		\includegraphics[width=\textwidth]{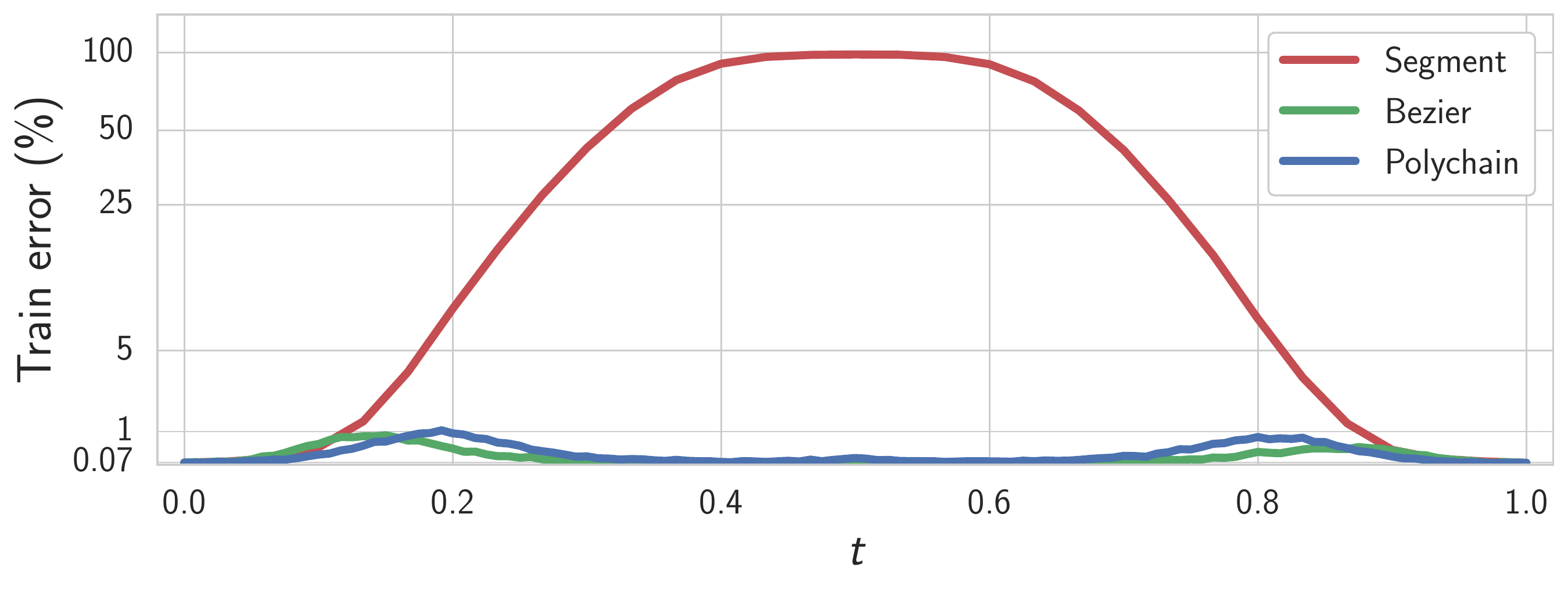}
	\end{subfigure}
	\qquad
	\begin{subfigure}{0.46\textwidth}
		\includegraphics[width=\textwidth]{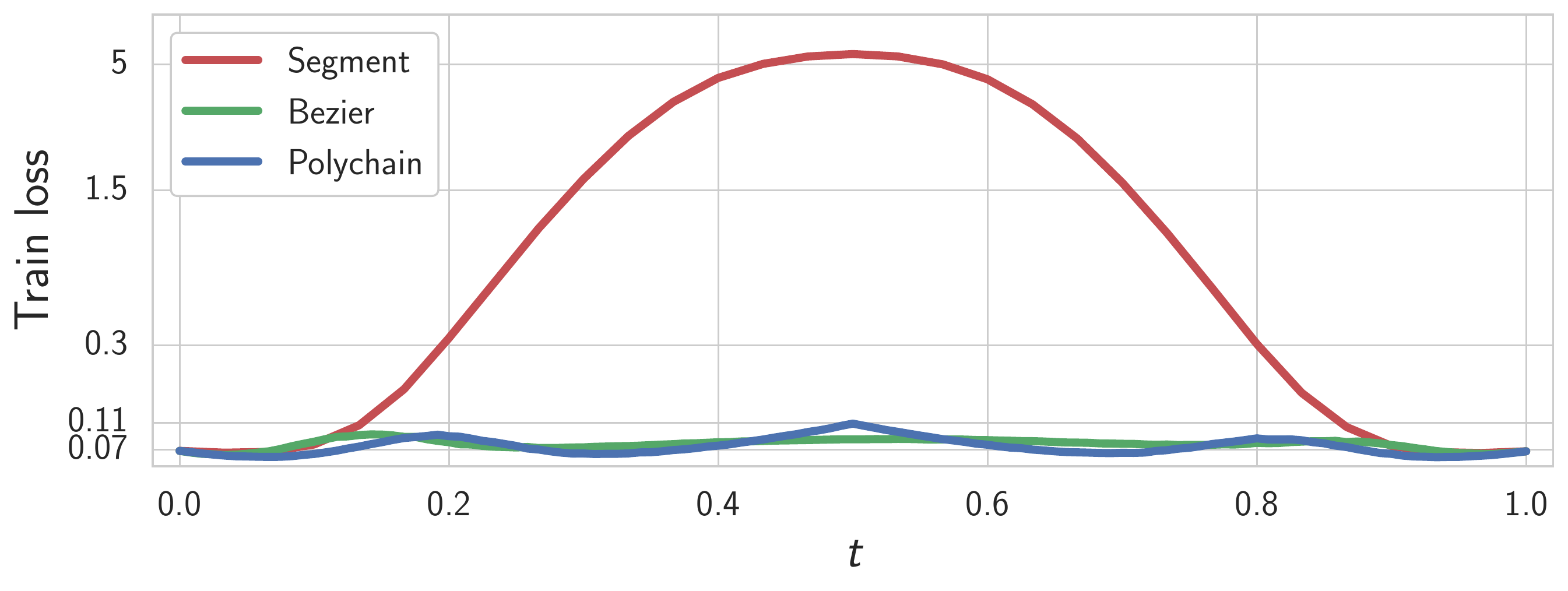}
	\end{subfigure}
	\\
	\begin{subfigure}{0.46\textwidth}
		\includegraphics[width=\textwidth]{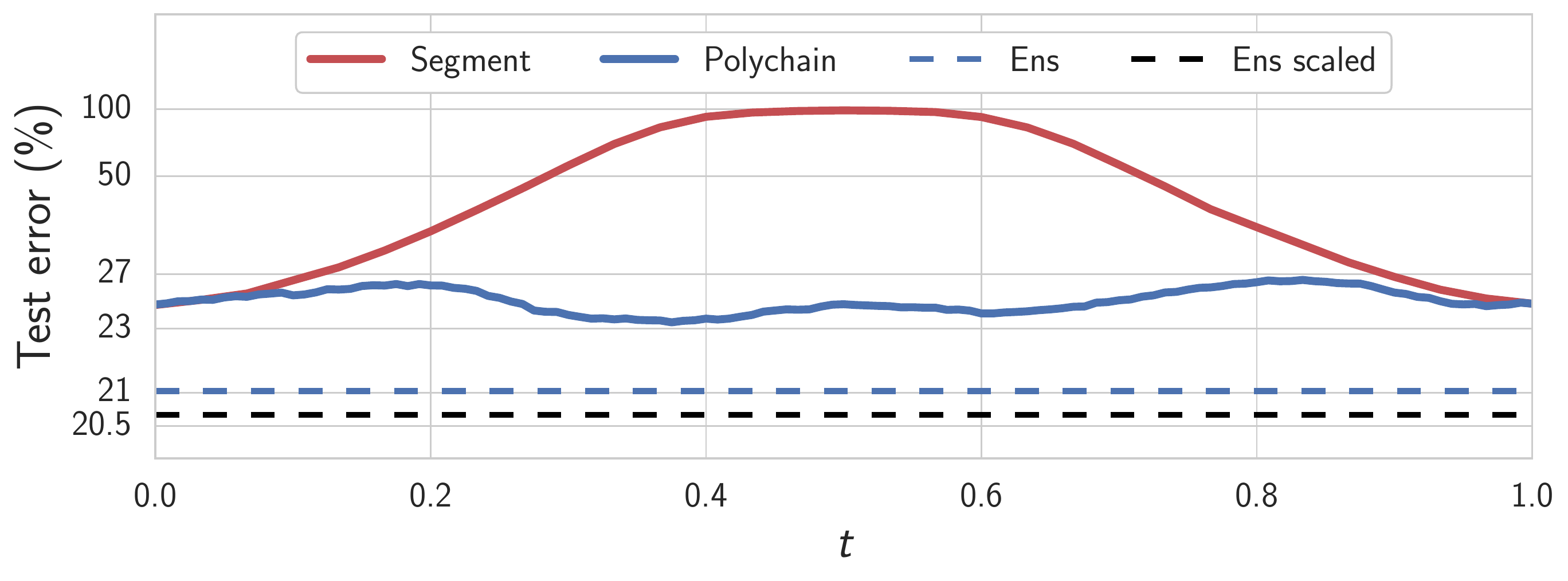}
	\end{subfigure}
	\qquad
	\begin{subfigure}{0.46\textwidth}
		\includegraphics[width=\textwidth]{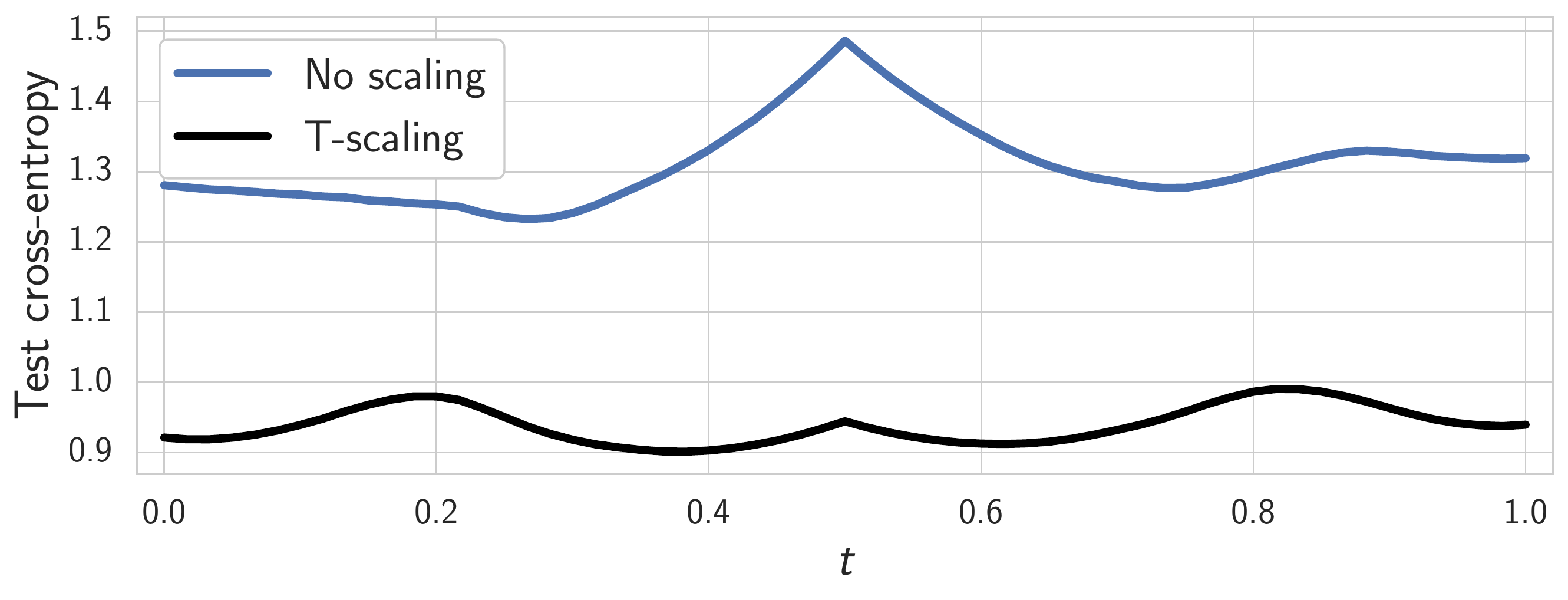}
	\end{subfigure}    
	\caption{Error as a function of the point on the curves $\phi_{\theta}(t)$ found
		by the proposed method, using a ResNet-164 on CIFAR-100. {\bfseries Top left:} train error.
		{\bfseries Bottom left:} test error; dashed lines correspond to quality of 
		ensemble constructed from curve points before and after logits rescaling.
		{\bfseries Top right:} train loss ($\ell_2$ regularized cross-entropy).
		{\bfseries Bottom right:} cross-entropy before and after logits rescaling for 
		the polygonal chain.}
	\label{fig:t_curves}
\end{figure*}

Here we explore ensembles constructed from points sampled from these high 
accuracy curves.
In particular, we train a polygonal chain with one bend connecting two independently trained 
ResNet-164
networks on CIFAR-100 and construct an ensemble of networks corresponding
to $50$ points placed on an equally-spaced grid on the curve. The resulting
ensemble had $21.03\%$ error-rate on the test dataset. The error-rate of the ensemble
constructed from the endpoints of the curve was $22.0\%$.  An ensemble of three
independently trained networks has an error rate of $21.01\%$.  
Thus, the ensemble of the networks on the curve outperformed an ensemble
of its endpoints implying that the curves found by the proposed method are
actually passing through diverse networks that produce predictions different
from those produced by the endpoints of the curve. Moreover, the ensemble based on 
the polygonal chain has the same number of parameters as three independent networks,
and comparable performance.

Furthermore, we can improve the ensemble on the chain without adding additional parameters 
or computational expense, by accounting for the pattern of increased training and
test loss towards the centres of the linear paths shown in Figure~\ref{fig:t_curves}.
While the training and test accuracy are relatively constant, the pattern of loss, 
shared across train and test sets,
indicates overconfidence away from the three points defining the curve: in this 
region, networks tend to output probabilities closer to $1$, sometimes with the 
wrong answers. This overconfidence decreases the performance of ensembles constructed from 
the networks sampled on the curves. In order to correct for this overconfidence and 
improve the ensembling performance we use temperature scaling \citep{guo2017}, which is
inversely proportional to the loss. 
Figure \ref{fig:t_curves}, bottom right, illustrates the test loss of ResNet-164 on CIFAR-100
before and after temperature scaling.
After rescaling the predictions of
the networks, the test loss along the curve decreases and flattens. Further, the
test error-rate of the ensemble constructed from the points on the curve went down
from $21.03\%$ to $20.7\%$ after applying the temperature scaling, outperforming
$3$ independently trained networks.

However, directly ensembling on the curves requires manual intervention for 
temperature scaling, and an additional pass over the training data for each 
of the networks ($50$ in this case) at test time to perform batch normalization 
as described in section~\ref{sec:batchnorm}.  Moreover, we also need to train
at least two networks for the endpoints of the curve. 

\subsection{The Effects of Increasing Parametrization}
\label{sec:curve_understanding_exp}

\begin{figure}[!h]
	\centering
	\includegraphics[width=0.44\textwidth]{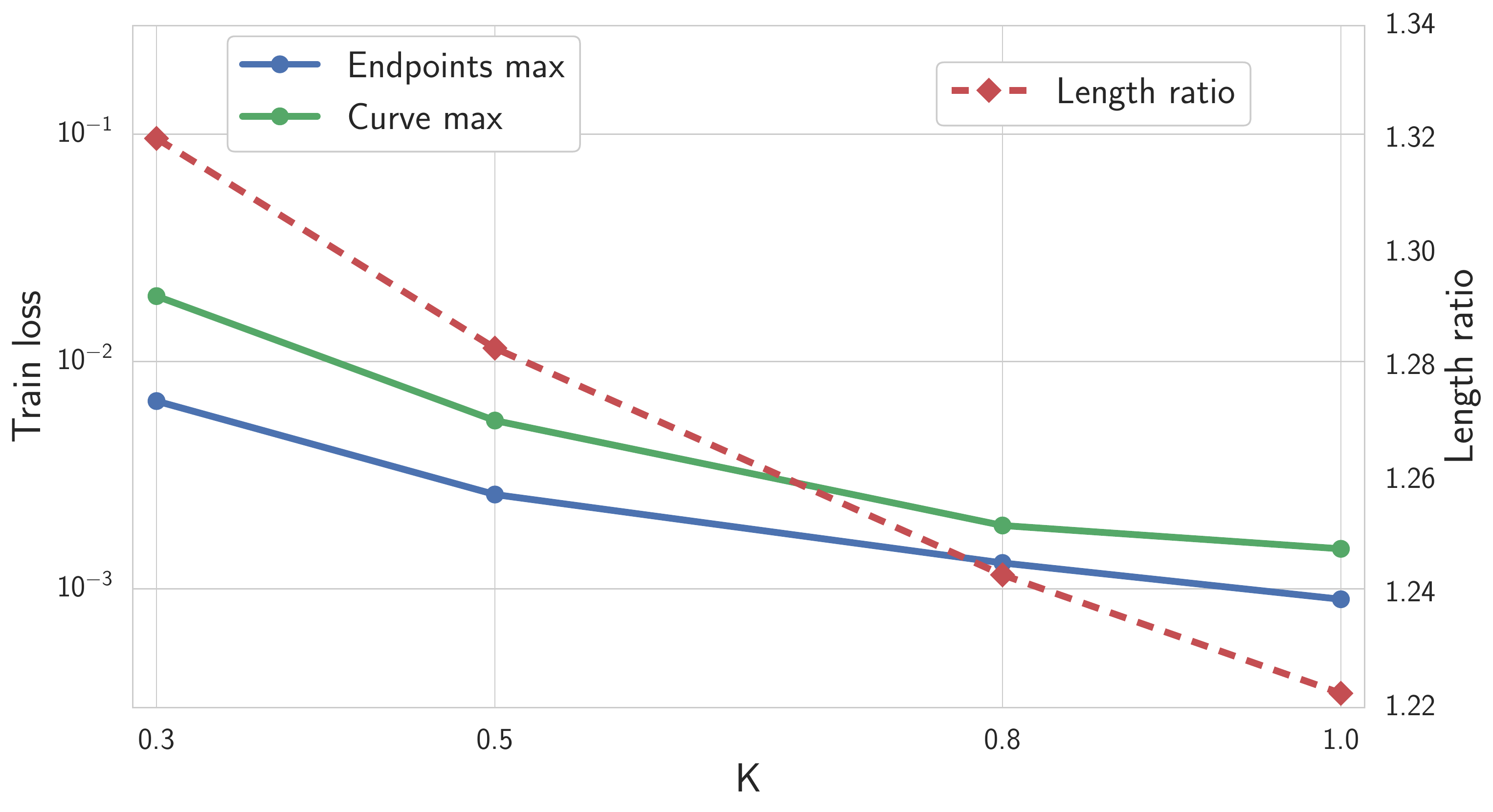}
	
	\caption{	
		The worst train loss along the curve, maximum of
		the losses of the endpoints, and the ratio of the length of the curve
		and the line segment connecting the two modes, as a function of
		the scaling factor $K$ of the sizes of fully-connected layers.
	}
	\label{fig:overparam}
\end{figure}

One possible factor that influences the connectedness of a local minima set is the overparameterization of neural networks. In this section, we investigate the relation between the observed connectedness
of the local optima and the number of parameters (weights) in the
neural network.  We 
start with a network that has three convolutional layers followed by
three fully-connected layers, where each layer has $1000K$ neurons.
We vary $K \in \{0.3, 0.5, 0.8, 1\}$, and for each value of $K$ we 
train two networks that we connect with a Bezier curve using 
the proposed procedure. 

For each value of $K$, Figure~\ref{fig:overparam} 
shows the worst training loss along the curve, maximum of losses
of the endpoints, and the ratio of the length of the curve
and the line segment connecting the two modes.  
Increasing the number of parameters we are able to reduce the 
difference between the worst value of the loss along the curve and
the loss of single models used as the endpoints. The ratio of the length of the found curve and the length of the line
segment connecting the two modes also decreases monotonically with $K$. 
This result is intuitive, since a greater parametrization allows for more flexibility in how we can 
navigate the loss surfaces.  

\subsection{Trivial connecting curves}
\label{sec:trivial_curves}

For convolutional 
networks with ReLU activations and without batch normalization we can construct 
a path connecting two points in weight space such that the accuracy of 
each point on the curve (excluding the origin
of the weight space)
is at least as good as 
the minimum of the accuracies of the endpoints. 
Unlike the paths found by our procedure, these paths are trivial 
and merely exploit redundancies in the parametrization. Also, the training loss
goes up substantially along these curves. Below we give a construction of such
paths.

Let $\hat w_1$ and $\hat w_2$ be two sets of weights.
This path of interest consists of two parts. The first part connects the point
$\hat w_1$ with $0$ and the second one connects the point $\hat w_2$ with  $0$. 
We describe  only the first part $\phi(t)$ of the path, such that 
$\phi(0) = 0, \phi(1) = \hat w_1$, as the second part is completely analogous. 
Let the weights of the network $\hat w_1$ be 
$\{W_i, b_i\}_{1 \leq i \leq n}$ where $W_i, b_i$ are the weights and biases of
the $i$-th layer, and $n$ is the total number of layers.
Throughout the derivation we consider the inputs of the network fixed.
The output of the $i$-th layer $o_i = W_i \mbox{ReLU}(o_{i-1}) + b_i$, $1 \leq i \leq n$, 
where $i = 0$ corresponds to the first layer and $i = n$ corresponds to logits
(the outputs of the last layer).
We construct $\phi(t) = \{W_i(t), b_i(t)\}_{1 \leq i \leq n}$ in the following 
way. We set $W_i(t) = W_it$ 
and $b_i(t) = b_i t^i.$ It is easy to see that logits of the network with 
weights $\phi(t)$ are equal to $o_n(t) = t^n o_n$ for all $t > 0$. 
Note that the predicted labels corresponding to the logits $o_n(t)$ and 
$o_n$ are the same, so the accuracy of all networks corresponding to $t > 0$
is the same.

\begin{algorithm}[!t]
	\caption{Fast Geometric Ensembling}
	\label{alg:FGE}
	\begin{algorithmic}
		\REQUIRE ~\\ weights $\hat w$, LR bounds $\alpha_1, \alpha_2$,\\ cycle length~$c$~(even), number of iterations $n$
		\ENSURE $\mbox{ensemble}$ 
		\STATE $w \leftarrow \hat w$ \COMMENT{Initialize weight with $\hat w$}
		\STATE $\mbox{ensemble} \leftarrow [~]$ 
		\FOR {$i \leftarrow 1, 2, \ldots, n$}
		\STATE $\alpha \leftarrow \alpha(i)$ \COMMENT{Calculate LR for the iteration}
		\STATE $w \leftarrow w - \alpha \nabla \mathcal{L}_i(w)$ \COMMENT{Stochastic gradient update}
		\IF{$\bmod(i, c) = c / 2$} 
		\STATE {$\mbox{ensemble} \leftarrow \mbox{ensemble} + [w]$} \COMMENT{Collect weights}
		\ENDIF
		\ENDFOR
	\end{algorithmic}
\end{algorithm}
\subsection{Fast geometric ensembling experiments}
\label{sec:fge_exp_details}
Alg.~\ref{alg:FGE} provides an outline of the algorithm.
As baseline models we used the following implementations:
\begin{itemize}
	\item VGG-16 (\url{https://github.com/pytorch/vision/blob/master/torchvision/models/vgg.py});
	\item Preactivation-ResNet-164 (\url{https://github.com/bearpaw/pytorch-classification/blob/master/models/cifar/preresnet.py});
	\item ResNet-50 ImageNet (\url{https://github.com/pytorch/vision/blob/master/torchvision/models/resnet.py});
	\item Wide ResNet-28-10 (\url{https://github.com/meliketoy/wide-resnet.pytorch/blob/master/networks/wide_resnet.py});
\end{itemize}

For the FGE (Fast Geometric Ensembling) strategy on 
ResNet we run the FGE routine summarized in Alg. $1$ after epoch $125$ of the 
usual (same as Ind) training for $22$ epochs. The total training time
is thus $125 + 22 = 147$ epochs.
For VGG and Wide ResNet models we run the pre-training procedure for $156$ epochs
to initialize FGE.
Then we run FGE for $22$ epochs starting from checkpoints corresponding to 
epochs $120$ and $156$ and ensemble all the gathered models. 
The total training time is thus $156 + 22 + 22 = 200$ epochs. 
For VGG we use cycle length $c = 2$ epochs, which
means that the total number of models in the final ensemble is $22$.
For ResNet and Wide ResNet we use $c = 4$ epochs, and the total number of models in the
final ensemble is $12$ for Wide ResNets and $6$ for ResNets.

\subsection{Polygonal chain connecting FGE proposals}
\label{sec:fge_curves}

\begin{figure}[!h]
	\centering
	\includegraphics[width=0.44\textwidth]{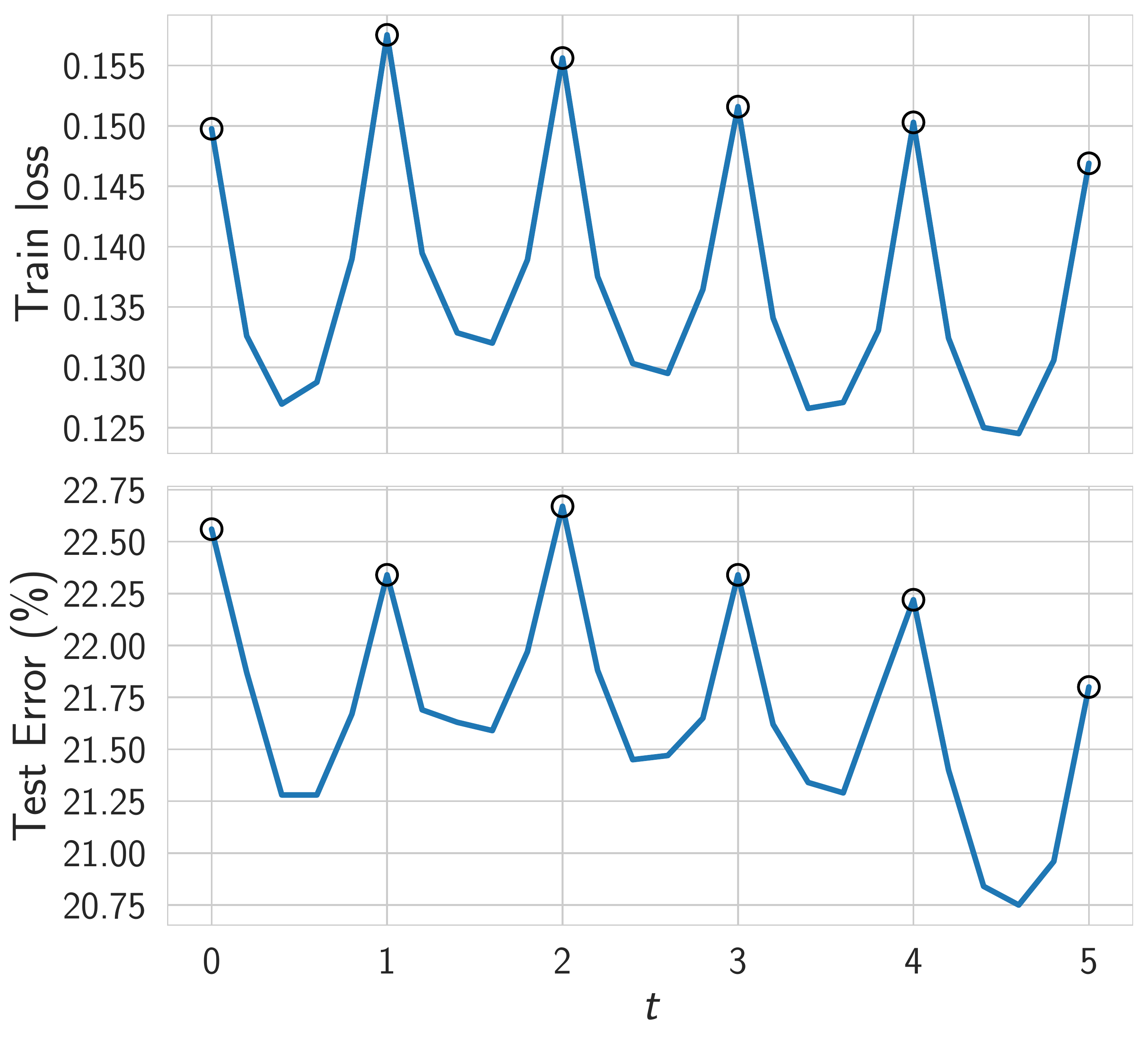}
	
	\caption{	
        Train loss and test error along the polygonal chain connecting the
        sequence of points ensembled in FGE. The plot is generated using 
        PreResNet-164 on CIFAR 100. Circles indicate the bends on the polygonal
        chain, i.e. the networks ensembled in FGE.
	}
	\label{fig:fge_polychain}
\end{figure}

In order to better understand the trajectories followed by FGE we construct
a polygonal chain connecting the points that FGE ensembles. Suppose we run 
FGE for $n$ learning rate cycles obtaining  $n$ points $w_1, w_2, \ldots, w_n$ 
in the weight space that correspond to the lowest values of the learning rate. 
We then consider the polygonal chain consisting of the line segments connecting 
$w_i$ to $w_{i+1}$ for $i=1, \ldots, n-1$. We plot test accuracy and train error
along this polygonal chain in Figure \ref{fig:fge_polychain}. We observe that
along this curve both train loss and test error remain low, agreeing with
our intuition that FGE follows the paths of low loss and error. Surprisingly,
we find that the points on the line segments connecting the weights $w_i, w_{i+1}$
have lower train loss and test error than $w_i$ and $w_{i+1}$. See 
\citet{izmailov2018averaging} for a detailed discussion of this phenomenon.

\end{document}